\documentclass[conference,usenames,dvipsnames]{IEEEtran}

\pdfoutput=1   

\usepackage{ifpdf}
\usepackage{booktabs}
\usepackage{multirow}
\usepackage[pagebackref=true,breaklinks=true,colorlinks,bookmarks=false]{hyperref}
\usepackage[dvipsnames]{xcolor}
\usepackage{graphicx}  
\usepackage[font=footnotesize,caption=false]{subfig}
\usepackage{caption} 

\usepackage{amsmath}
\usepackage{ulem}

\usepackage{booktabs, makecell, tabularx} 
\usepackage{rotating}
\usepackage[export]{adjustbox}

\usepackage{footnote}
\makesavenoteenv{tabular} 
\makesavenoteenv{table} 

\usepackage{flushend}  

\ifCLASSINFOpdf
\else
\fi

\IEEEoverridecommandlockouts 

\begin{document}
%
\title{SCNet: A Generalized Attention-based Model for Crack Fault Segmentation$^*$\thanks{$^*$To appear in ICVGIP'21}}

\author{\IEEEauthorblockN{Hrishikesh Sharma}
\IEEEauthorblockA{\textit{TCS Research} \\
hrishikesh.sharma@tcs.com}
\and
\IEEEauthorblockN{Prakhar Pradhan}
\IEEEauthorblockA{\textit{TCS Research} \\
prakhar.pradhan@tcs.com}
\and
\IEEEauthorblockN{Balamuralidhar Purushothaman}
\IEEEauthorblockA{\textit{TCS Research} \\
balamurali.p@tcs.com}
}


\maketitle

\begin{abstract}
Anomaly detection and localization is an important vision problem, having multiple applications. Effective and generic semantic segmentation of anomalous regions on various different surfaces, where most anomalous regions inherently do not have any obvious pattern, is still under active research. 
Periodic health monitoring and fault (anomaly) detection in vast
infrastructures, which is an important safety-related task, is one such
application area of vision-based anomaly segmentation. However, the task is
quite challenging due to large variations in surface faults, texture-less
construction material/background, lighting conditions etc. Cracks are
critical and frequent surface faults that manifest as extreme zigzag-shaped
thin, elongated regions. They are among the hardest faults to detect, even
with deep learning. In this work, we address an open aspect of automatic
crack segmentation problem, that of generalizing and improving the
performance of segmentation across a variety of scenarios, by modeling the
problem differently. We carefully study and abstract the sub-problems
involved and solve them in a broader context, making our solution generic.
On a variety of datasets related to surveillance of different
infrastructures, under varying conditions, our model consistently
outperforms the state-of-the-art algorithms by a significant margin,
without any bells-and-whistles. This performance advantage easily carried over in two deployments of our model, tested against industry-provided datasets. Even further, we could establish our model's performance for two manufacturing quality inspection scenarios as well, where the defect types are not just crack equivalents, but much more and different. Hence we hope that our model is indeed a truly generic defect segmentation model.
\end{abstract}


\IEEEpeerreviewmaketitle

\section{Introduction}
\label{intro_sec}

\IEEEPARstart{S}{emantic} Segmentation of anomalous regions on objects of
interest is an important vision task, with multiple industrial
applications. One such application area is related to infrastructures.
\textit{Infrastructures} are facilities and services set up and operated in
various nations, to support their economy, industry and living conditions.
They are vast man-made objects, whose examples include bridges, dams, rail
corridors, tunnels and pipelines, pavements and roads, towers and buildings etc. To maintain their safety, durability and other health parameters during their service life span, it is required that any fault that is developing over time is detected and attended to, via repairs, before it manifests into a widespread catastrophe. Hence \textit{periodic health inspections} of infrastructures, which can detect and quantify faults in a timely manner, are carried out. These inspections provide aid in prioritizing and planning of maintenance tasks.

There are multi-fold challenges that arise during health inspections. During remote visual surveillance for inspection, due to hazardous and unreachable premises through which infrastructures mostly pass, drones are being increasingly used to perform such surveillance. This mode of surveillance leads to capturing of large volumes of visual data, leading to increasing demand of automated visual analysis. Further, due to safety requirements, the visual data is captured from at least five meters of distance. This leads to nascent fault regions on the infrastructures being imaged as small-sized regions quite often, making their automated segmentation a tough task. The other challenges include types and variations of faults themselves. Many intrinsic variations in various fault regions are in fact \textit{continuous} factors of variations, e.g. region's shape, region's texture etc. For example, both efflorescence and cracks grow into very random shapes. Extrinsic factors of variations mainly arise from outdoor conditions such as lighting condition, surface wetness, occlusion and shadow effect etc. Due to such variations, the fault regions generally display a lack of any obvious region-wide pattern. This is perhaps the biggest challenge in automating visual fault detection.

\begin{figure*}
\begin{center}
\subfloat{\label{an_1}\includegraphics[scale=.29]{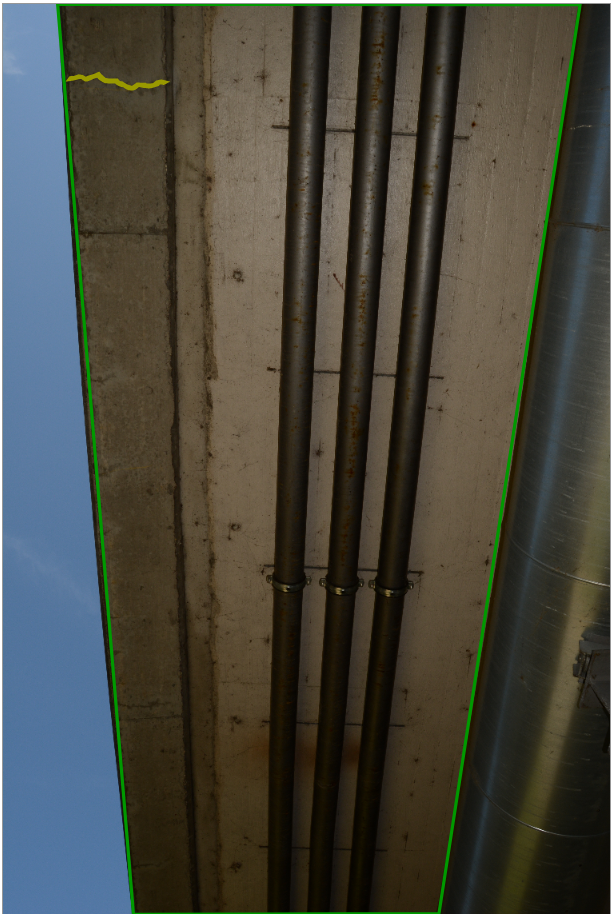}}
\qquad\qquad
\subfloat{\label{an_2}\includegraphics[scale=.29]{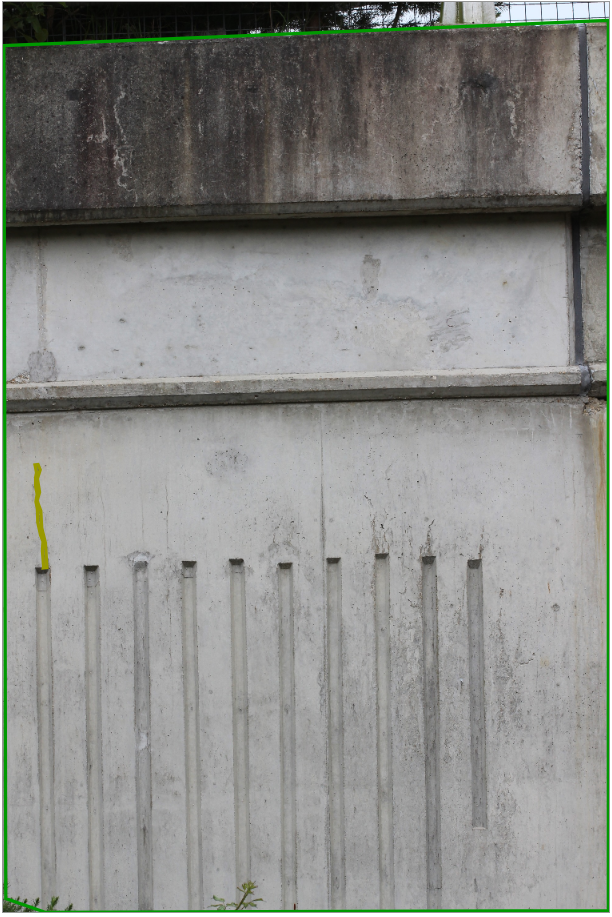}}
\caption{Example Crack Regions in Infrastructures, annotated with \textbf{\textcolor{GreenYellow}{Yellow}}-colored pixels}
\label{crack_ex_fig}
\end{center}
\end{figure*}

In terms of impact on structural health, detection of \textit{crack} fault
category is known to be a top priority maintenance task. Visually, cracks
being extremely thin, highly irregular and long, embedded at arbitrary
locations in backgrounds which represent diverse materials and hence
textures, they are among hardest to detect in an image (refer
Fig.~\ref{crack_ex_fig}). Cracks are known to co-occur along with other
faults, such as efflorescence \cite{codebrim_pap}, making the
classification task multi-target and thus exacerbating their analysis.
Given its importance and complexity, a lot of prior research on crack
detection has been carried out, in past few decades. Till few years back,
crack detection was carried out mainly based on \textit{intensity-based
methods}. These methods had the problem of requiring to manually tune
various parameters such as thresholds to deal with different lighting
conditions etc. Hence the applicability and reliability of these methods
was limited. In recent years, deep learning (DL-)based methods operate by
making task-specific decisions by looking at a larger context
hierarchically. Hence they have demonstrated much better performance in
computer vision as well as many other domains. The DL-based methods for
crack detection broadly fall in two categories. \textit{Semantic
Segmentation}(pixel-level classification) approach is preferred over
patch-based classification for crack detection, because it enables robust
downstream task of crack mechanical analyses. Such downstream analyses can
be further used to predict crack growth over time
\cite{crack_structural_pap, crack_meas_pap}. Our literature study showed
that the prior semantic segmentation methods have been mostly biased towards usage of a single crack dataset, more often being a road/pavement dataset, as is also corroborated in \cite{dc_orig_pap}. Their affinity towards one dataset introduces implicit bias, leading to overfitting that limits their effectiveness. In another words, given multiple factors of variations inherent in faults of various nature as described earlier, these methods are unable to cope with the \textit{domain shift} between different datasets.

In this paper, we take a different approach towards the problem of semantic
segmentation of crack regions. We find out important \textbf{generic}
sub-problems, which occur not only in crack segmentation task, but also in
other non-crack vision tasks, sometimes occurring beyond even semantic
segmentation. Some examples include, as will be explained in section
\ref{model_sec}, usage of attention to compensate for lack of contextual
information, dealing with class imbalance etc. When we solve these
sub-problems in a \textit{broader} context, we are able to get the required
generalizability across surveillance scenarios, as well as \textit{consistent}
improvement in performance, in terms of F1 score and area under P-R
curve(AUPRC). We prove our approach's effectiveness and portability over multiple datasets, arising out of multiple industrial contexts and different infrastructures, and as a bonus, two manufacturing scenarios as well.

To summarize, our proposed method has the following contributions:
\begin{itemize}
    \item We design a new soft attention mechanism, to be used in the multi-scale feature space.
    \item We carefully use a novel multi-task combination of binary focal
loss and softIoU loss to simultaneously handle class imbalance sub-problem as well as reduce the pixel-level shift in the prediction.
    \item We generate and additionally input a noisy \textit{prior} about crack region, which improves performance.
\end{itemize}

The rest of this paper is organized as follows. We summarize the related prior research in section \ref{lit_surv_sec}. In section \ref{model_sec}, we present design details of SCNet model, as well as solutions to the broader sub-problems within. It is followed by overview of datasets used in this paper, in section \ref{dataset_sec}. The results and performance of SCNet is presented in section \ref{res_sec}. We present a detailed ablation study to highlight the importance of various design factors in section \ref{abl_sec}, before we conclude the paper.

\section{Related Work}
\label{lit_surv_sec}
Here, we provide brief overview of recent advances in research on topics that are relevant to our problem.

\subsection{Vision-based Crack Detection}
Crack detection has been the mainstay of research on visual surveillance of faults in infrastructures, over last many decades. Its applicability has been researched upon in variety of application scenarios: tunnel safety inspection \cite{tunnel_crack_pap}, \cite{rt_tunnel_crack_pap}, dam safety inspection \cite{dam_crack_pap}, rail corridor inspection \cite{track_insp_pap}, steel bridges \cite{weld_crack_pap}, \cite{girder_crack_pap}, concrete bridges \cite{codebrim_pap}, mine safety \cite{ground_crack_pap}, pavement crack detection \cite{dc_orig_pap}, \cite{deep_2_pap}, building safety inspection \cite{bldg_defect_pap}, to name a few. The techniques for detection belong to two phases. For prior methods which do not employ deep learning, see comprehensive literature survey in \cite{infra_review_pap}. While very recently sporadic deep generative modeling based approaches have started appearing \cite{crackgan_pap}, almost all the deep learning based approaches in last few years have been supervised discriminative approaches, for obvious reasons. These DL-based discriminative approaches can further be grouped into two important categories.

\subsubsection{Classification-based Approaches}
The initial DL-based approaches were patch-level classification approaches \cite{concrete_ptch_pap}, \cite{ccr_pap}, \cite{rcd_dcnn_pap}, \cite{crack_class_pap}, \cite{nb_cnn_pap},\cite{ptch_cd_pap}, \cite{mason_class_pap}, \cite{patch_wsl_pap}, due to smaller datasets. Usage of patches instead of whole images allowed having more training data. However, the performance of these methods was average.

\subsubsection{Segmentation-based Approaches}
More recently, with appearance of larger crack datasets on variety of surfaces, the research community has focused upon more fine-grained detection task, via semantic segmentation \cite{crack_pattern_pap}, \cite{crack_fcn_pap}, \cite{pav_crack_pap}, \cite{ground_crack_pap}, \cite{rcd_rsf_pap}, \cite{opt_crack_pap}. As mentioned earlier, we too follow this approach, and design to get a versatile solution.

\subsection{Semantic Segmentation}
Starting from earlier important works such as SegNet and FCN, there have been many important advances on the task of semantic segmentation. Usage of multi-scale context \cite{scale_attn_pap} and careful architecture design are main concerns of segmentation. Many architectures resemble an autoencoder-style hourglass architecture, with the recent U-Net \cite{unet_pap} being quite popular. As motivated in FCN, all recent important architectures are fully convolutional, and use residual propagation. For a more in-depth survey of recent segmentation approaches, see \cite{semseg_survey_pap}.

The main limitation of these approaches is that they are designed for
segmentation of non-anomalous scenes such as CityScapes, Kitti, organs in
biomedical images etc. As is known, CNNs are biased towards learning
texture \cite{texture_bias_pap}, while surfaces of infrastructures are
mostly texture-less e.g. steel, concrete etc. \cite{texture_study_pap}.
Hence the naive applicability of established methods is quite poor, as we
also found experimentally. 

\subsection{Attention Mechanism}
Attention mechanism is an attempt to mimic human brain's ability to
selectively concentrate on relevant objects, while ignoring rest of the
scene and the objects. At few times, such relevant objects relate to
anomalies: our brain is able to effortlessly localize an abnormal
aspect/region within a scene. The recent breakthrough modeling of attention
happened in NLP area, wherein different parts of contextual information are
selectively weighted using learnt weights, to attend and highlight the
(current) input token. This mechanism was first adapted in computer vision in \cite{show_attend_tell_pap}, with weighting over various regions of background/context. This became the foundation of most of the future attention-based solutions to vision tasks. 
Recently, attention mechanism was also adopted in various semantic segmentation approaches \cite{local_attn_pap}, \cite{ccnet_pap}, \cite{psanet_pap}, \cite{scale_attn_pap}. The approach in \cite{psanet_pap} is closest to ours approach in solving one sub-problem. However, while they focus on learning pixel-specific attention masks based on spatial distance in input image, to implement non-local attention, we focus on learning attention masks in the pyramidical feature space (scale-specific attention), for a reason described in section \ref{attn_sec}. This focus is also the reason behind our work being more precise than \cite{attn_unet_pap}, \cite{pave_pan_pap}, \cite{mixed_attn_pap}, where attention modules are used in the residual paths.

\section{SCNet Model}
\label{model_sec}

We model the problem as a segmentation of foreground crack region vs. rest
of background region. We take a \textbf{Supervised} approach, rather than an unsupervised approach, since it is indicated in a path-breaking work \cite{unsup_dis_pap} that whenever sufficient ground truth annotations are available, supervised approaches are still likely to outperform unsupervised approaches on downstream tasks. 
The architecture part of our
model is an extension of the well-known and well-performing (supervised) UNet
segmentation model \cite{unet_pap}. 
Much like instance segmentation, where generation of foreground mask is a
primary concern, here our concern is to focus upon and predict a crack mask
only. Unlike instance segmentation, where we found that the poor concurrent prediction of crack bounding boxes, which are very foreground-sparse, pulls down the performance of mask generation as well, in a multi-task learning setup, here we focus on pixel-level prediction via a \textit{semantic segmentation} task. In the \textbf{S}upervised \textbf{C}rack\textbf{Net} (SCNet) design, we solve the following important sub-problems carefully, not addressed in the baseline or elsewhere.

\begin{figure*}
\begin{center}
\subfloat[Image]{\label{r_1}\includegraphics[scale=.3]{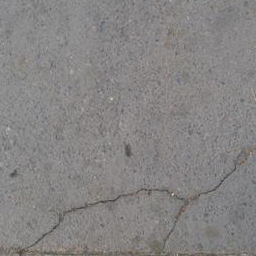}}
\quad
\subfloat[Ground-Truth]{\label{r_2}\includegraphics[scale=.3]{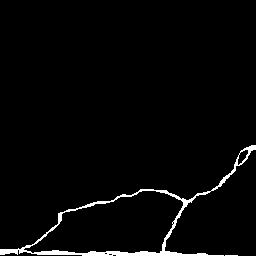}}
\quad
\subfloat[Prediction]{\label{r_3}\includegraphics[scale=.3]{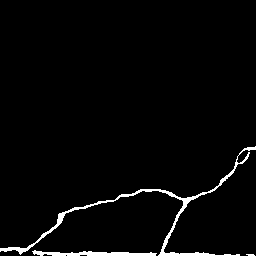}}
\quad
\subfloat[Overlap]{\label{r_4}\includegraphics[scale=.3]{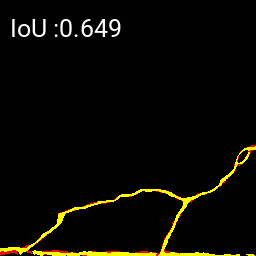}}
\quad
\subfloat[Zoom view of Overlap]{\label{r_5}\includegraphics[scale=.3]{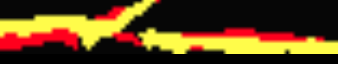}}
\caption{Disproportional degradation of IoU measure w.r.t. minor shift in predicted segmentation. Yellow: Predicted Crack Pixels; Red: GT Crack Pixels}
\label{displaced_fig}
\end{center}
\end{figure*}

\subsection{Discriminative Crack Representation}
\label{attn_sec}
Cracks, like many other anomalies, are \textit{local phenomenon}, occurring
randomly on a surface at any location. Hence there is no useful global cue
for their localization, in the pixel space. Instead of such cue, we turn to
attention mechanism to enhance the likelihood of a pixel being classified
into a foreground pixel. Our novel attention module enables the encoder 
to learn a foreground representation, which is more discriminative towards surface faults including cracks.

It was brought out in FCN that different stages of the convolutional layers
within the encoder bring out diverse information about the objects. More
specifically, \cite{cnn_repr_pap} showed that first layer learns edge-like
artefacts, second layer learns contours and corners, and so on. Thus, the
features of lower layers mostly represent structure information, while the
features of deeper layers mostly represent shape/semantic information of a
region. Cracks being thin regions with random shapes which absorb light and
are dark, any useful representation of them has to weigh more on the
structural aspect of it. Hence we employ \textit{attention operator in the
scale-space} naturally implied within the feature pyramid of an encoder, to
implement such relative weighting. This relative weighting is applicable to
segmentation of cracks irrespective of the surface on which they appear,
lending it the required \textit{generality}. 

\begin{figure}[!h]
\begin{center}
\includegraphics[scale=.16]{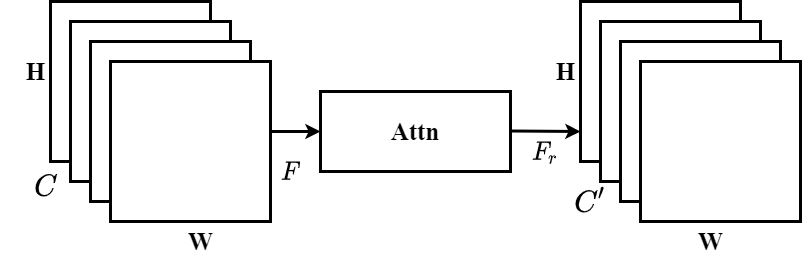}
\includegraphics[scale=.15]{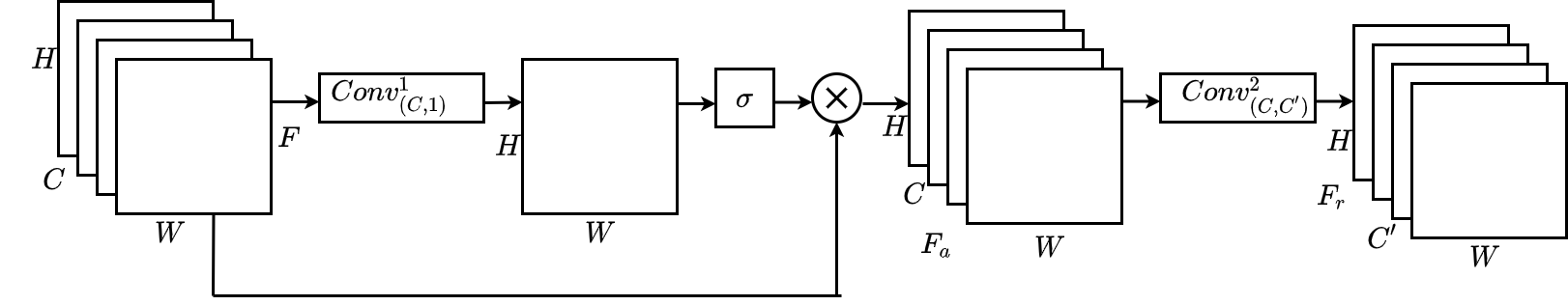}
\caption{Scale-wise Attention Module Structure}
\label{attn_mod_fig}
\end{center}
\end{figure}

Our design of scale-space attention operator works convolutionally over a
block of feature maps, $F$, of size $H\times W \times C$. As shown in Fig.~\ref{attn_mod_fig}, in this design,
we do not use any pooling mechanism. The intuition is that the foreground
being highly imbalanced as compared to background, there is not much
information in the feature maps, and any kind of pooling will further
trivialize the feature information contained in the feature maps, leading
to decreased performance. We could corroborate this using experiments
later. The attention masks so generated are multiplied with the original
feature maps, $F$, to get intermediate feature maps, $F_a$, as per general
practice. We experimentally found that using an additional convolutional
layer, $Conv^{2}_{(C,C^\prime)}$ over $F_a$, gives more refined feature maps, $F_r$ which further enhance the performance of
the model. 
To summarize,

\begin{equation}
\label{attn_eq}
\begin{split}
F_{a} = \sigma\left(Conv^{1}_{(C,1)}\left(F\right)\right)\times F \\
F_{r} = Conv^{2}_{(C,C^\prime)}\left(F_{a}\right)
\end{split}
\end{equation}
where $\mathbf{\sigma}$ is the sigmoid activation function, $\mathbf{\times}$ denotes the element-wise multiplication and $Conv_{(m,n)}$ denotes a convolutional layer with $m$ number of input channels and $n$ number of output channels. We embed our novel attention operator in the feature pyramid, in the same way as \cite{cbam_pap}.

\subsection{Minimizing FPs from Reconstruction-based blur}
Using attention only in the scale space of the encoder improves the true
positives, but leads to many pixels being misclassified as false
positives(FPs), especially near ground truth (GT) boundaries. Our interpretation is that this happens due to well-known problem of additional blur in autoencoder-style reconstructions \cite{ae_blur_pap} (our architecture resembles autoencoder-style hourglass architecture). Cracks being highly imbalanced foregrounds, ignoring false positives leads to drastic fall in segmentation performance. To ameliorate this situation, we symmetrically use attention modules in decoder as well. Such usage indeed led to significant drop in amount of pixel-wise false positives, once again across a diverse set of surfaces.

\subsection{Reducing Impact of Class Imbalance}
\label{model_sec_focal}
Cracks being thin, elongated objects (sometimes 3-5 pixels wide
\cite{dam_crack_pap}), the ratio of crack pixels vs. non-cracks pixels
(refer Table~\ref{data_imb}) is quite low. This leads to \textit{severe} class imbalance in training data. Learning effective classifiers in presence of class imbalance, which do not trivially misclassify foreground into background, is a long-researched topic in machine learning. Among all known ways to tackle \textit{severe} class imbalance, we employ a pixel-wise adaptation of focal loss to learn an effective classifier. \textbf{Focal loss} was recently proposed in context of object detection task, to train a bounding box classifier in presence of severe imbalance \cite{focal_loss_pap}, but its design is generic. As ablation studies show, usage of focal loss in place of binary cross-entropy loss \textit{with} median frequency balancing, systematically reduced the negative impact of imbalance across all datasets. We also tried to use another recent loss-based approach to tackle imbalance, the Dice loss \cite{dice_loss_pap}, but usage of focal loss was found much superior to this loss.

\begin{figure*}
\begin{center}
\includegraphics[scale=.19]{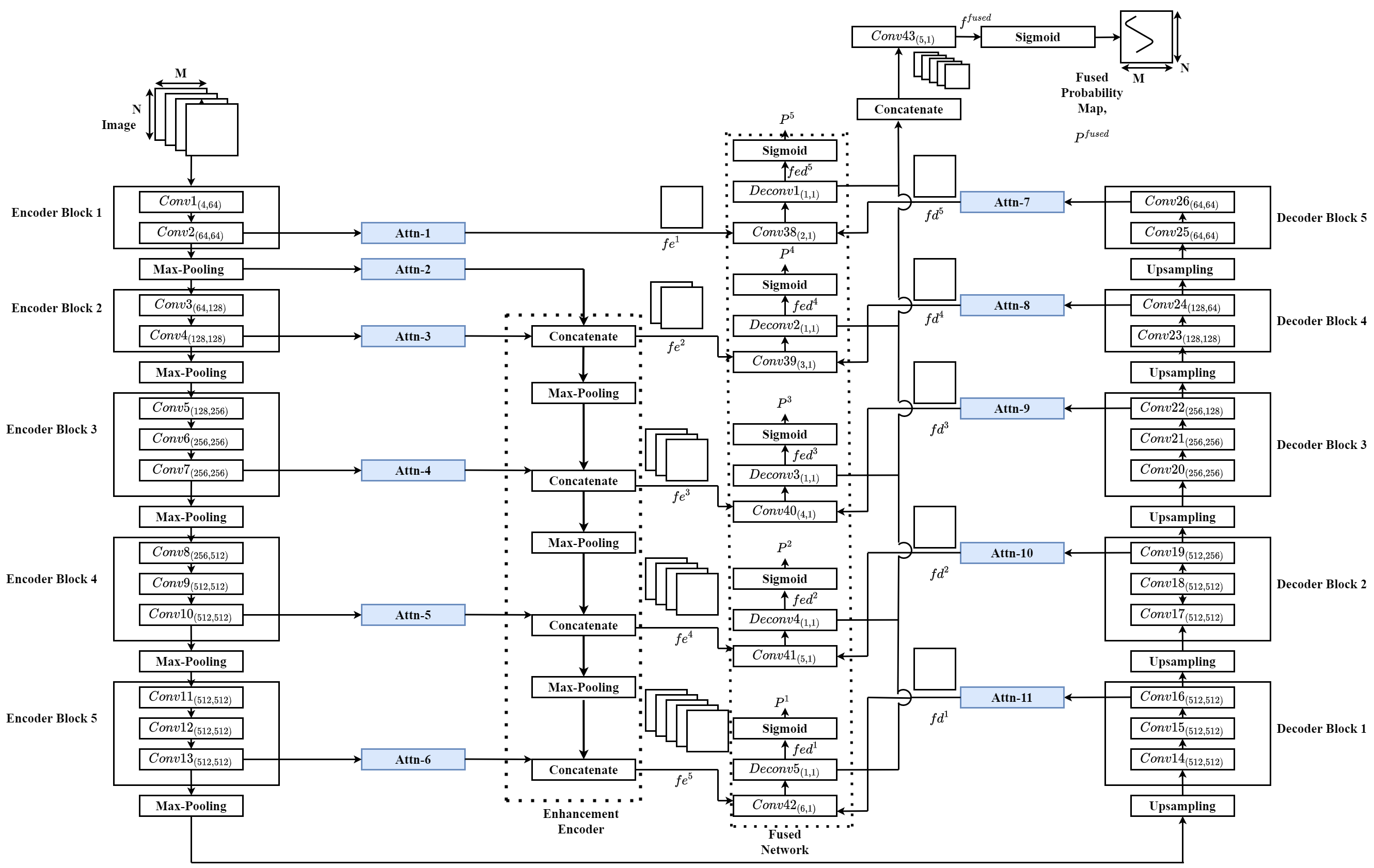}
\caption{SCNet Architecture}
\label{scnet_arch_fig}
\end{center}
\end{figure*}

\subsection{Tighter Alignment of Predictions to GT Regions}
\label{model_sec_soft}
We empirically found out one more reason for limited performance of
existing crack segmentation models: poor localization. As is shown in
Fig.~\ref{displaced_fig}, even after using attention and handling class
imbalance, the predicted crack region was found to be offset by few pixels
against the ground truth (GT). Cracks being very thin regions, few pixels'
shift, especially in the direction perpendicular to its approximate major axis of its
axis-aligned bounding box, leads to non-trivial fall in overlap, a
localization measure. The most popular overlap measure is
intersection-over-union(IoU), but is non-differentiable. To reduce the
above fall, we employed and tested various approximations of IoU as an
auxiliary loss, designed
for usage in semantic segmentation task. Specifically, we implemented and
used Lov\'asz loss \cite{lovasz_loss_pap}, Soft-IoU loss
\cite{softiou_loss_pap} and NeuroIoU loss \cite{neuro_iou_pap}. As is shown
in Table~\ref{loss_tab}, usage of Soft-IoU loss gave us relatively better
performance improvement. We then used soft-IoU loss in conjunction with
pixel-wise focal loss, in a \textit{multi-task learning} setup, as in
Eq.~\ref{total_loss}. As is shown in \cite{mtl_uncertain_pap}, not all tasks can work advantageously together, in a MTL setup. However, it is known from research over object detection that a classification and a regression loss supplement each other quite well, in box prediction task. We found the same synergistic effect of these losses in pixel-level prediction task as well.

\subsection{Biasing with a Prior}
Since cracks resemble morphologically dilated edges of various shapes, we
also experimented with biasing the input with a location prior, using
edges. 
 Specifically, we used the popular HED algorithm \cite{hed_pap} to generate a noisy edge map. We then use the edge map in a different way, by adding it as an additional channel to the RGB input. Introduction of additional channel led to marginal improvement in performance, across various datasets.

\subsection{Overall Architecture}
The proposed architecture is built on an extension of UNet called HCNN
\cite{hcnn_pap}, and hence has an hourglass architecture. Such structure
results in somewhat sharper reconstruction of object boundaries
\cite{pan_net_pap}. The network, as shown in Fig.~\ref{scnet_arch_fig},
consists of $4$ sub-networks, attention based encoder, attention-based
decoder, enhancement encoder and a fused network. 

The \textbf{encoder} follows the VGG-C \cite{vgg_16} network architecture, containing $5$ encoder blocks with $13$ convolutional layers placed sequentially. Usage of max pooling instead of dilated convolution helps us to retain the max pooling indices, and transfer subsequently to the decoder, helping to recover the loss of details in the representation space.
The feature maps output by each encoder block are further passed through
the attention modules(Fig. \ref{attn_mod_fig}), to focus on the crack regions within them.

The \textbf{enhancement encoder} is used to improve the feature maps from the deeper encoder blocks. The purpose of performing iterative fusion of feature maps from successive encoder blocks, within this sub-network, is explained in \cite{hcnn_pap}.

The attention-driven \textbf{decoder} is used in the network, for reconstructing the crack region from focused feature maps. It performs bilinear interpolation-driven upsampling to decode the region from the latent feature space representation. Specifically, the upsampling incorporates the pooling indices transferred by the encoder, to perform \textit{non-linear} upsampling.

In the \textbf{fused} sub-network, feature maps from the enhancement encoder and decoder at the corresponding scale are fused using a $1 \times 1$ convolutional module. The output of the convolutional module is then up-sampled to the input image size and passed through sigmoid activation to calculate the predicted probability maps at each stage. There are thus $5$ predicted probability maps. To obtain a final fused probability map, the feature maps, $f_{ed}^{i}$ from each scale are concatenated and passed through a convolutional module and sigmoid activation function. The probability map so obtained, is passed through an iterative thresholding prcoess \cite{dc_orig_pap} to obtain a binary crack probability map. In the complete network, the feature maps are concatenated along the channel dimension.

The entire SCNet model, and the sub-networks within, can be mathematically described by a system of equations. For one example sub-network, e.g. fused network, the equations that represent the generation of the final fused probability map are shown as follows.
\begin{equation}
\begin{split}
    f_{ed}^{5} = Deconv1_{(1,1)}(Conv38_{(2,1)}(Concat(f_{e}^{1},f_{d}^{5})) \\
    f_{ed}^{4} = Deconv2_{(1,1)}(Conv39_{(3,1)}(Concat(f_{e}^{2},f_{d}^{4})) \\
    f_{ed}^{3} = Deconv3_{(1,1)}(Conv40_{(4,1)}(Concat(f_{e}^{3},f_{d}^{3})) \\
    f_{ed}^{2} = Deconv4_{(1,1)}(Conv41_{(5,1)}(Concat(f_{e}^{4},f_{d}^{2})) \\
    f_{ed}^{1} = Deconv5_{(1,1)}(Conv42_{(6,1)}(Concat(f_{e}^{5},f_{d}^{1}))
    \end{split}
\end{equation}

Each encoder block, and units in other block at same horizontal level, together signify $i^{th}$ scale. Let $P^{i}$ denote the pixel-wise probability map obtained at $i^{th}$ scale in the fused sub-network, and $P^{fused}$ denote the final fused probability map. Then,

\begin{equation}
    P^{i} = \frac{1}{1+e^{-f_{ed}^{i}}}
\end{equation}
\begin{equation}
    f^{fused} = Conv43_{(5,1)}(Concat(f_{ed}^{5},f_{ed}^{4},f_{ed}^{3},f_{ed}^{2},f_{ed}^{1}))
\end{equation}
\begin{equation}
    P^{fused} = \frac{1}{1+e^{-f^{fused}}}
\end{equation}
where $Conv_{(m,n)}$ denotes a convolutional layer, with $m$ number of input channels, and $n$ number of output channels.

\subsection{Loss Formulation}
To handle per-pixel classification with class imbalance, along with perceived shift in predicted region, we use a novel combination of binary focal loss  and Soft-IoU loss in a \textit{multi-task learning} setup.
Further, to enhance the probability of true classification, contribution of predictions from the individual scales is also factored in the overall loss, a form of \textit{deep supervision}.

Let the training dataset consist of $K$ $\langle$image, ground-truth$\rangle$ pairs, $\left\{(X^{k},Y^{k}),k = 1,2,\ldots K\right\}$, where, $Y^{k}=\left\{y_j^k,j=1,2,\ldots M\times N, y_j^k\in\left\{0,1\right\}\right\}$. Let $L^{focal}_{i}$ denote the focal loss at the $i^{th}$ stage, computed using $f_{ed}^{i}$. Also, let $L^{focal}_{fused}$ denote the focal loss calculated using the $f^{fused}$. Similarly, let $L^{IoU}$ denote the Soft-IoU loss calculated using the fused probability map $P^{fused}$. Then, the overall loss can be computed as
\begin{equation}
\label{total_loss}
    L^{total}_{focal} = L^{fused}_{focal} + \sum_{i=1}^5w_{i}*L^{i}_{focal}
\end{equation}
\begin{equation}
    L^{i}_{focal}=\sum_{j=1}^{MN}\ell_{1}(f_j^{i},y_{j})
\end{equation}
\begin{equation}
    L^{fused}_{focal}=\sum_{j=1}^{MN}\ell_{1}(f_j^{fused},y_{j})
\end{equation}
\begin{equation}
    L^{total}_{focal} = \sum_{j=1}^{MN}\ell_{1}(f_j^{fused},y_{j}) + \sum_{j=1}^{MN}\sum_{i=1}^5w_{i}*\ell_{1}(f_j^{i},y_{j})
\end{equation}
\begin{equation}
    L^{IoU} = \ell_{2}(P^{fused},Y)
\end{equation}
\begin{equation}
\label{eq_7}
    L_{total} = L^{total}_{focal} + L^{IoU}
\end{equation}
where $w_{i}$ are scalar weights used to give relative importance, \(\ell_{1}\) is focal loss function defined in \cite{focal_loss_pap} and \(\ell_{2}\) is Soft-IoU loss function defined in \cite{softiou_loss_pap}.



\section{Datasets}
\label{dataset_sec}
As specified earlier, we have targeted our model design for versatility as
well as acceptably high performance. To verify the effectiveness of design,
we have carefully chosen a group of datasets to benchmark the performance.
The choice is driven by the consideration of their \textit{heterogeneity
and size}. We have also taken different deployment scenarios into consideration: aerial/drone-based as well as terrestrial, again to prove effectiveness across various deployment scenarios. Some primary statistics about the datasets are as in Table~\ref{dset_tab}. For few datasets, which did not provide an explicit test set, we held out $20\%$ of images for testing the model. The individual details of each dataset can be found in the corresponding papers, cited within first column. Two datasets arising out of industrial deployment scenarios are listed in last two rows, and do not have any corresponding papers.


\begin{table}[!h]
\begin{center}
\caption{Summary of Public Datasets Used}
\label{dset_tab}
\setlength{\tabcolsep}{0.5em} 
{\renewcommand{\arraystretch}{1.2}
\begin{tabular}{|c|c|c|}
\hline
\textbf{Name} & \textbf{Infrastructure} & \textbf{Size (crack samples)} \\ \hline \hline
Deepcrack \cite{dc_orig_pap} & Pavements  & 1953 \\ \hline
CODEBRIM\footnote{Additional region-wise annotations are available from \cite{codebrim_extra_ann}} \cite{codebrim_pap} & Bridges & 2281 \\     \hline
METU \cite{metu_dset} & Building Facade & 12115 \\ \hline
CrackForest \cite{rcd_rsf_pap} & Road Images & 323 \\ \hline
Wessex Water & Water Pipeline Inner Walls & 4874 \\ \hline
BMW & Road Images & 5892 \\ \hline
\end{tabular}
}
\end{center}
\end{table}

\section{Experiments and Results on Public Datasets}
In this section, we demonstrate our goal of achieving versatility and improvement in performance, by a set of carefully designed experiments. We test our model from both semantic segmentation and anomaly detection point of views, since crack is also a surface anomaly and occurs as a rare spatial event.
We have tested our model without any bells-and-whistles, especially
CRF-based post-processing that is commonly used in certain semantic segmentation
networks.

\label{res_sec}
\subsection{Implementation}
We implemented SCNet model using PyTorch. 
 All the weights and biases have been initialized using Xavier initialization. We used SGD \cite{sgd_pap} as the optimization technique for the model training. The size of the input images is $256 \times 256$, while the mini-batch size used is $4$. We used an initial learning rate of $0.0001$, with momentum value of $0.9$, and weight decay of $2e-4$. The relative weights assigned to both focal loss and Soft-IoU loss in \ref{eq_7} is $1.0$. The $\alpha$ and $\gamma$ used in the focal loss are $1.0$ and $2.0$ respectively. To further enhance the training process, the RGB pixel values of the input image is normalized between $-1$ and $+1$. The scalar weights used in the equation~\ref{total_loss} to give relative importance in the final prediction  are $\langle 0.5, 0.75, 1.0, 0.75, 0.5\rangle$ respectively. For the attention mechanism in eq. ~\ref{attn_eq}, we empirically found $C^\prime=1$ giving best performance. The input image to the network is a $4$ channel image, where the fourth channel is a binary edge map of the RGB image. To compute the binary edge map, we used a public implementation of HED \cite{pytorch-hed}.
 
\subsection{Data Augmentation}
To achieve fair comparison with the baseline \cite{hcnn_pap}, we followed
similar data augmentation process as used in \cite{hcnn_pap}, rotating the
images from $0$ to $90$ degrees, flipping the image horizontally and
vertically, randomly cropping the flipped images with a size of $256 \times
256$. 
The percentage of crack and non-crack pixels for each dataset after data augmentation is quoted in the Table~\ref{data_imb}, which shows the existence of severe data imbalance problem.
 
\begin{table}[!h]
\begin{center}
\caption{Degree of Class Imbalance in Datasets}
\label{data_imb}
\setlength{\tabcolsep}{0.5em} 
{\renewcommand{\arraystretch}{1.2}
\begin{tabular}{|c|c|c|}
\hline
\textbf{Name} & \textbf{Crack Pixels $\left(\%\right)$} & \textbf{Non-crack Pixels $\left(\%\right)$} \\ \hline \hline
Deepcrack \cite{dc_orig_pap} & 4.68 & 95.31 \\ \hline
CODEBRIM  \cite{codebrim_pap} & 3.78 & 96.21 \\ \hline
METU \cite{metu_dset} & 7.02 & 89.47 \\ \hline
CrackForest \cite{rcd_rsf_pap} & 3.04 & 96.95 \\ \hline
Wessex Water & 6.78  & 93.22 \\ \hline
BMW & 11.63 & 88.37
\\ \hline
\end{tabular}
}
\end{center}
\end{table}


\begin{figure*}[!t]
\begin{center}
\subfloat[PRC for CODEBRIM]{\label{pr_code}\includegraphics[width=.23\textwidth]{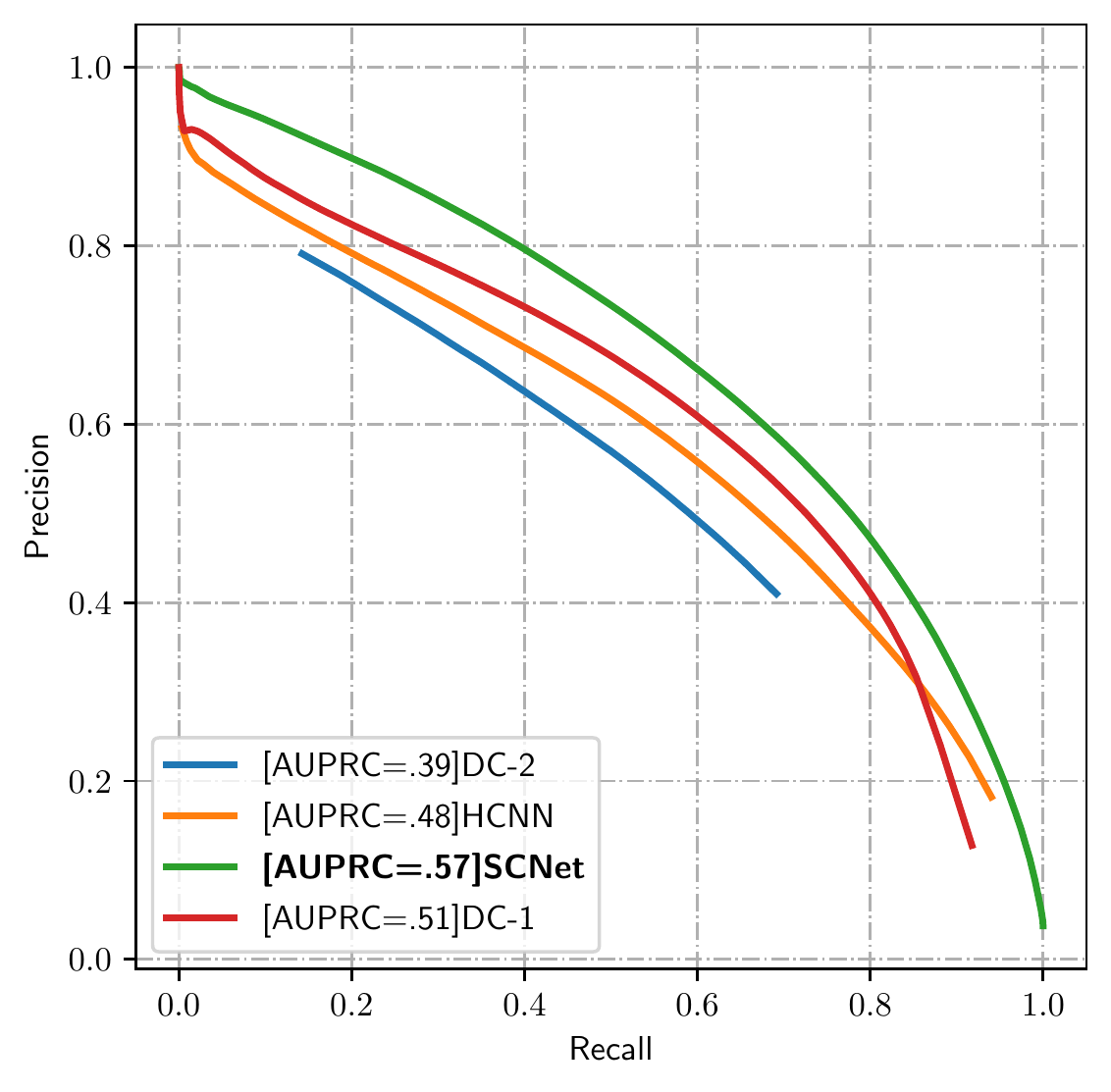}}
\quad
\subfloat[PRC for CrackForest]{\label{pr_cfd}\includegraphics[width=.23\textwidth]{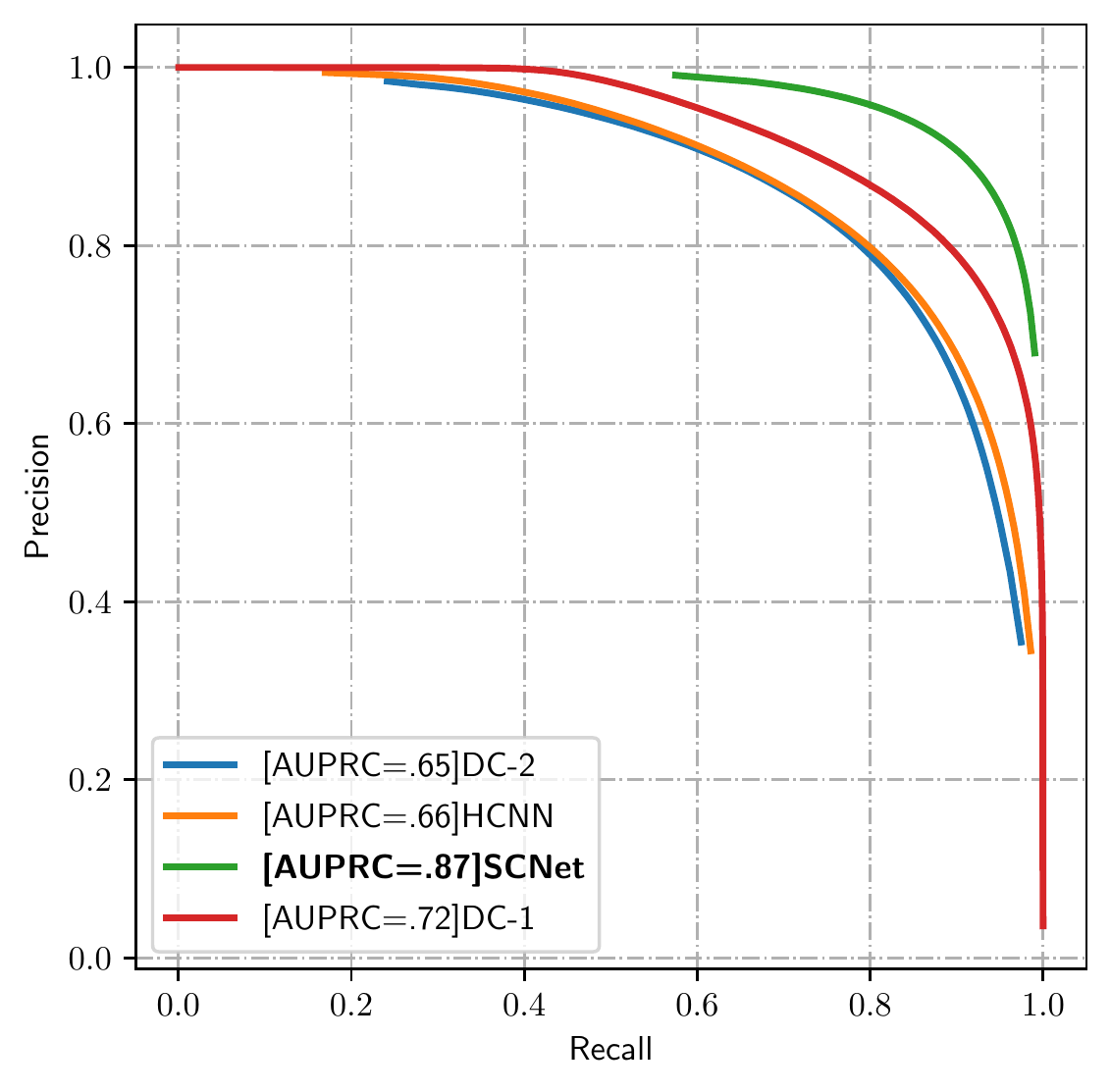}}
\quad
\subfloat[PRC for DeepCrack]{\label{pr_deep}\includegraphics[width=.23\textwidth]{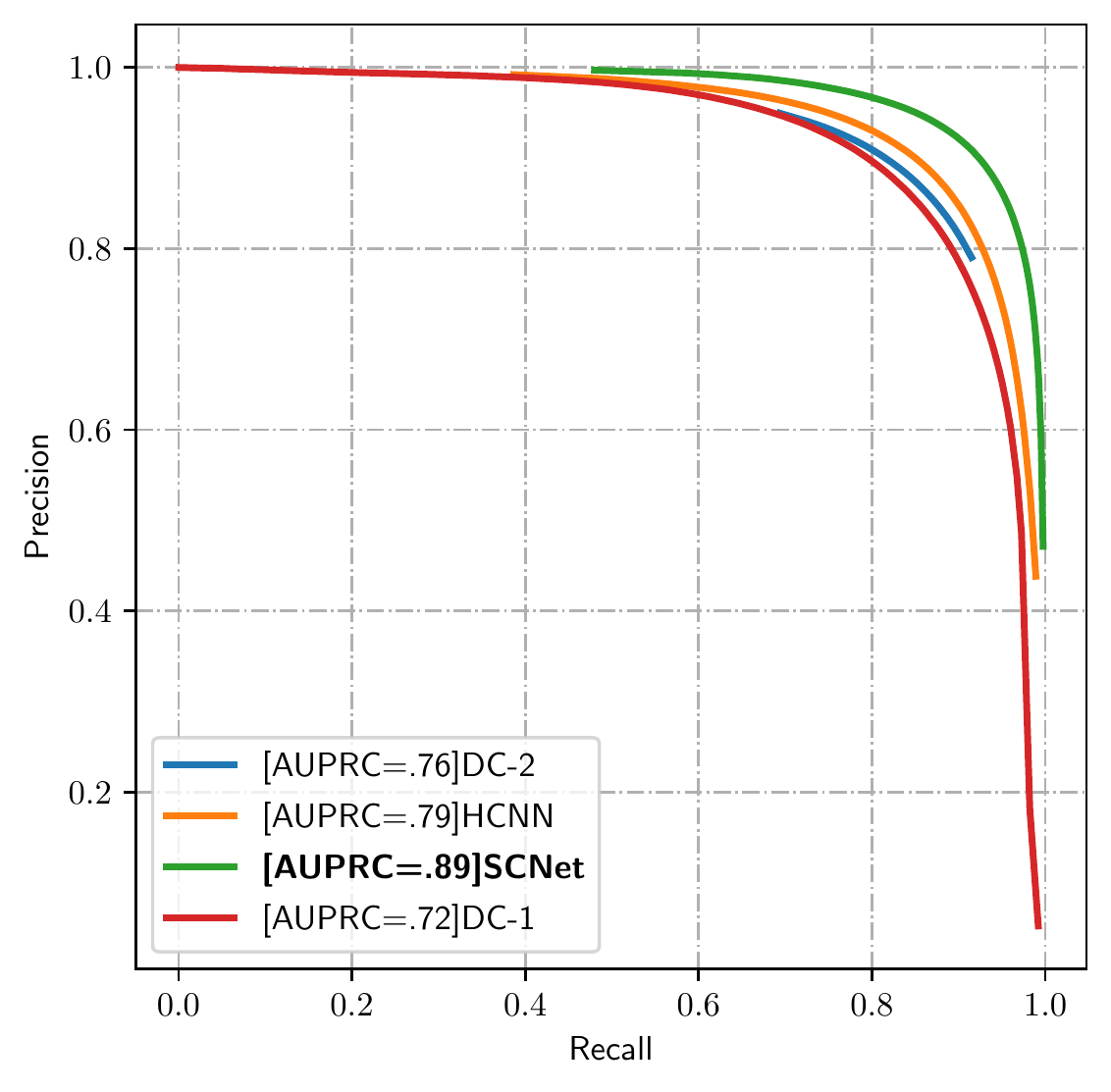}}
\quad
\subfloat[PRC for METU]{\label{pr_metu}\includegraphics[width=.23\textwidth]{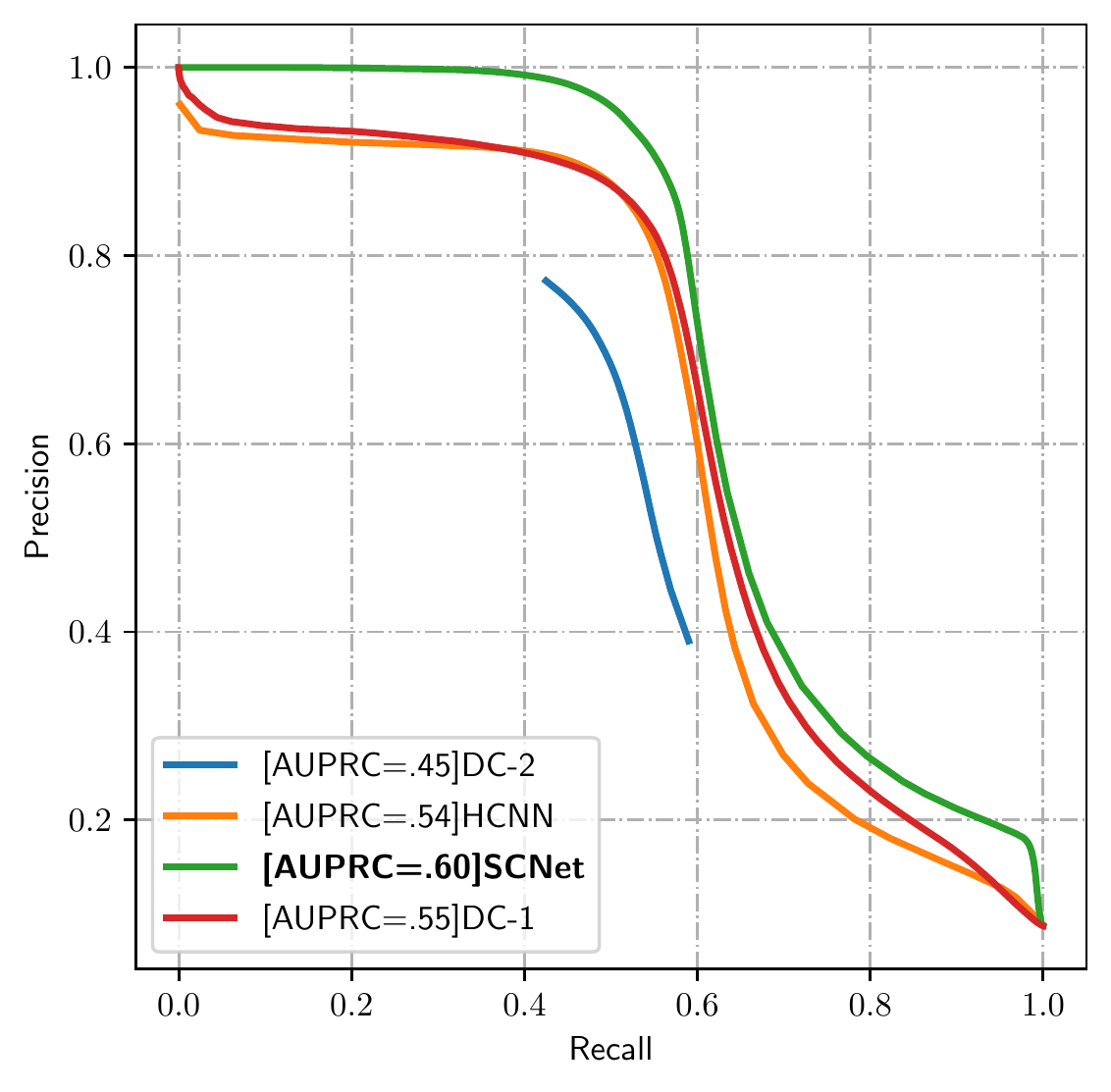}}
\caption{P-R Curve and AUPRC of various models, on each dataset}
\label{pr_auprc_fig}
\end{center}
\end{figure*}

\subsection{Evaluation Metric}
We report the pixel-wise \textbf{F1 score} of the crack class, as the primary evaluation metric. Both the false positives and true negatives are identified with respect to crack pixels only, since we are interested in segmentation of the lone foreground class. To compute this score, we use an iterative thresholding process \cite{dc_orig_pap} over $P^{fused}$. The iterative thresholding can be seen as computing the maximum F1 score via \textit{finite} sampling of the corresponding P-R Curve. A very similar metric which considers both FPs and TNs is IoU of the foreground class. Though we only report the more popular figure in literature on this problem, F1 score, we also measured the IoU metric. The trends in IoU performance are \textbf{fully consistent} in the sense that on all datasets, SCNet performs the best by a fair margin. From another angle, though crack is an anomaly, and area under Receiver Operating Curve (ROC) is a popular metric in anomaly detection, for binary classification under extreme imbalance scenarios, PRC is used instead of ROC for reasons described in \cite{prc_pap}. Hence we report both PRC and F1 score in next section.

There are circumstances when one has to evaluate and compare against both patch-based and pixel-based semantic segmentation models, as in \cite{mix_eval_pap}. Where we have to similarly compare against patch-based classification, we approximately convert our pixel-wise metric into \textit{region-wise F1 score} of $32\times 32$-sized patches. In a ground-truth image, $X$, a patch is said to have a crack, when the number of crack pixels in the patch constitutes at least $5\%$ of the total pixels (number based on statistics in Table~\ref{data_imb}). For the predicted  probability map, a region is said to have a crack, when the model can correctly detect at least $50\%$ of the crack pixels. Then the region-wise F1 score can be calculated by finding out true positive, false positive and false negative patches.

\begin{equation}
    Precision_{patch-wise}=\frac{TP_{\#patches}}{TP_{\#patches}+FP_{\#patches}}
\end{equation}

\begin{equation}
    Recall_{patch-wise}=\frac{TP_{\#patches}}{TP_{\#patches}+FN_{\#patches}}
\end{equation}
\begin{equation}
    F1_{patch-wise}=\frac{2\times Prec_{patch-wise}\times Rec_{patch-wise}}{Prec_{patch-wise} + Rec_{patch-wise}}
\end{equation}






\begin{table}[!h]
\begin{center}
\caption{Comparative Pixel-wise F1 Scores}
\label{pf1_tab}
\setlength{\tabcolsep}{0.5em} 
{\renewcommand{\arraystretch}{1.2}
\begin{tabular}{|c|c|c|c|c|}
\hline
\multirow{2}{*}{\textbf{Dataset}} & \textbf{DC-1\cite{dc_orig_pap}} & \textbf{HCNN} & \textbf{DC-2\cite{deep_2_pap}} & \textbf{SCNet} \\%
 & & \textbf{(Baseline)} & & \textbf{(Ours)} \\ \hline \hline
Deepcrack & 85.2 & 87.6 & 86.2 & \textbf{91.23} \\ \hline
CODEBRIM & 60.7 & 57.9 & 54.2 & \textbf{64.22} \\ \hline
METU & 66.10 & 65.40  & 57.80 & \textbf{69.13} \\ \hline
CrackForest & 84.42 & 79.91 & 79.58 & \textbf{90.40} \\ \hline
\end{tabular}
}
\end{center}
\end{table}

\subsection{Quantitative Performance}
For each shortlisted dataset, we have chosen to compare against
those state-of-the-art models, whose reported performance could be
reproduced. In some cases, we had to do our own implementation of such
models carefully\footnote{The baseline HCNN model was reimplemented by us,
since code was unavailable.}. Another concern was that the evaluation
metric used by various papers is quite varied, so not all works were
comparable on our evaluation metric of choice\footnote{Prominent omission
for this reason is MetaQNN model used by CODEBRIM}. It so turns out that we
could reproduce results having the chosen F1 score metric, of three models.
Two of these models are \textit{natively} associated with two shortlisted
datasets, and are introduced in the corresponding (dataset) paper. Hence,
we used these three \textit{recent} models, and along with our model, re-trained and
evaluated their performace on all four datasets. The F1 scores are shown in Tables~\ref{pf1_tab} and \ref{rf1_tab}.
It can be clearly observed from these tables that our model outperforms
other models, on \textbf{all} datasets, in terms of F1 score. The \textit{minimum
advantage} on pixel-wise F1 score is $\mathbf{4.09\%}$(on Deepcrack dataset), though in general it gives much higher advantage on other datasets and deployment scenarios. On patch-wise F1 score, our model outperforms state-of-the-art by quite a significant margin.

\begin{table}[!h]
\begin{center}
\caption{Comparative Region-wise F1 Scores}
\label{rf1_tab}
\setlength{\tabcolsep}{0.5em} 
{\renewcommand{\arraystretch}{1.2}
\begin{tabular}{|c|c|c|} \hline
\label{reg_f}
\textbf{Dataset} & \textbf{Patch-based Classification} & \textbf{SCNet(ours)} \\ \hline \hline
Deepcrack & 62.53 & \textbf{97.03}  \\ \hline
CODEBRIM & 38.67 & \textbf{74.69} \\ \hline
METU & 47.74 & \textbf{77.48} \\ \hline
CrackForest & 66.84 & \textbf{96.62} \\ \hline
\end{tabular}
}
\end{center}
\end{table}

\begin{figure*}
\setlength\tabcolsep{1pt}
\settowidth\rotheadsize{CrackForest}
\begin{center}
\begin{tabularx}{.85\linewidth}{l XXXXX}
\rothead{DeepCrack}   &   \includegraphics[width=\hsize,valign=m]{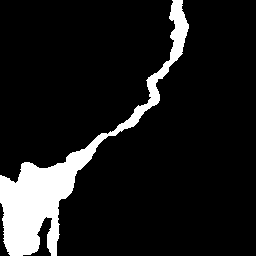}
                       &  \includegraphics[width=\hsize,valign=m]{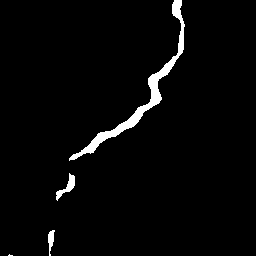} 
                       &  \includegraphics[width=\hsize,valign=m]{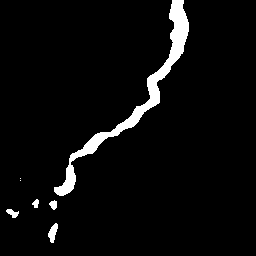}    
                       &   \includegraphics[width=\hsize,valign=m]{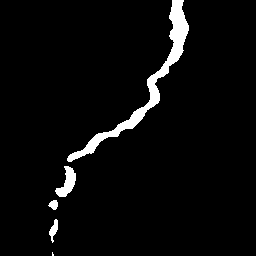}
                       &  \includegraphics[width=\hsize,valign=m]{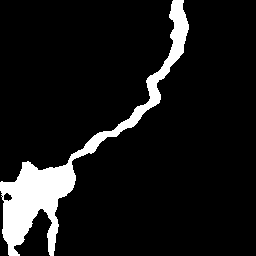}  \\  \addlinespace[4pt]
\rothead{CrackForest}   &   \includegraphics[width=\hsize,valign=m]{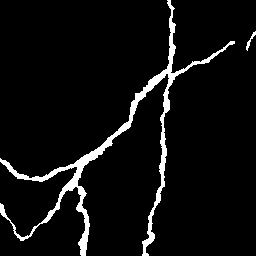}
                       &  \includegraphics[width=\hsize,valign=m]{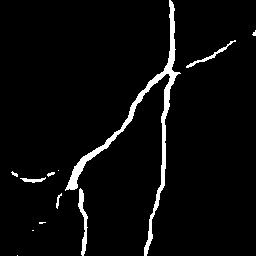} 
                       &  \includegraphics[width=\hsize,valign=m]{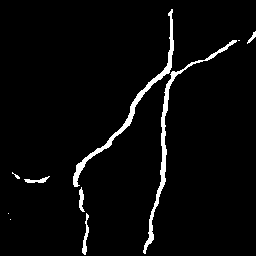}    
                       &   \includegraphics[width=\hsize,valign=m]{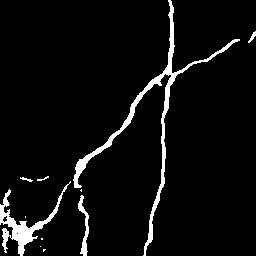}
                       &  \includegraphics[width=\hsize,valign=m]{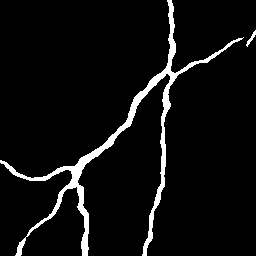}  \\  \addlinespace[4pt]
\rothead{CODEBRIM}      &   \includegraphics[width=\hsize,valign=m]{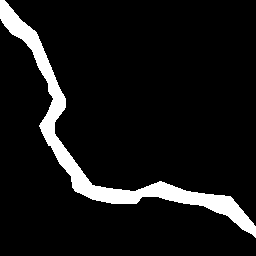}
                       &   \includegraphics[width=\hsize,valign=m]{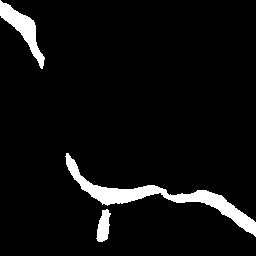}
                       &   \includegraphics[width=\hsize,valign=m]{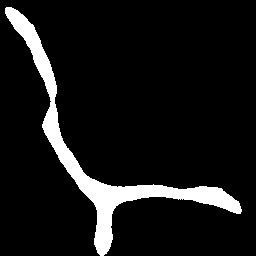}    
                       &   \includegraphics[width=\hsize,valign=m]{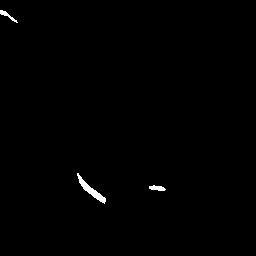}
                       &   \includegraphics[width=\hsize,valign=m]{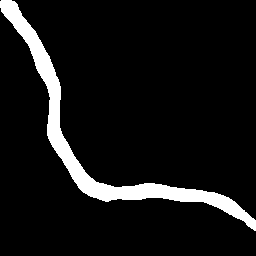} \\  \addlinespace[4pt]
\rothead{METU} &   \includegraphics[width=\hsize,valign=m]{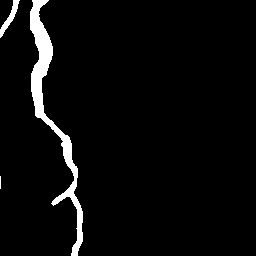}
                       &   \includegraphics[width=\hsize,valign=m]{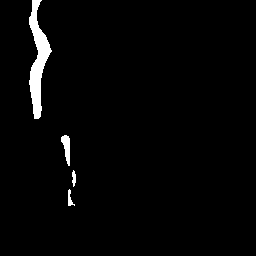}
                       &   \includegraphics[width=\hsize,valign=m]{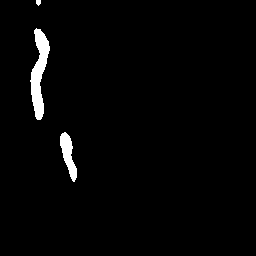}    
                       &   \includegraphics[width=\hsize,valign=m]{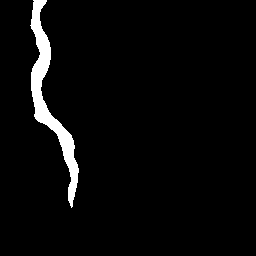}
                       &   \includegraphics[width=\hsize,valign=m]{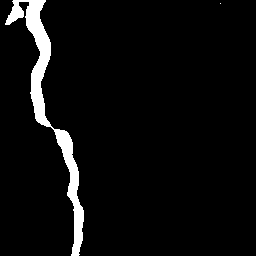} \\  \addlinespace[4pt]
                       & \centering Ground Truth & \centering DC-1 & \centering HCNN & \centering DC-2 & \centering Ours
\end{tabularx}
\end{center}
\caption{Qualitative Comparison of Segmentation Results, across Models and Datasets}
\label{seg_res_fig}
\end{figure*}

\begin{figure*}[!h]
\begin{center}
\subfloat[Scale-1,$P^{1}$]{\label{prob_1}\includegraphics[width=.18\textwidth]{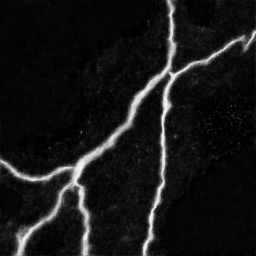}}
\quad
\subfloat[Scale-2,$P^{2}$]{\label{prob_2}\includegraphics[width=.18\textwidth]{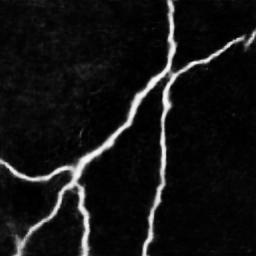}}
\quad
\subfloat[Scale-3,$P^{3}$]{\label{prob_3}\includegraphics[width=.18\textwidth]{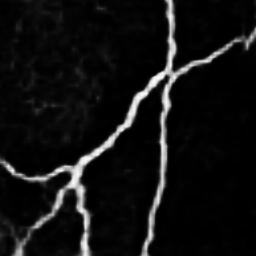}}
\quad
\subfloat[Scale-4,$P^{4}$]{\label{prob_4}\includegraphics[width=.18\textwidth]{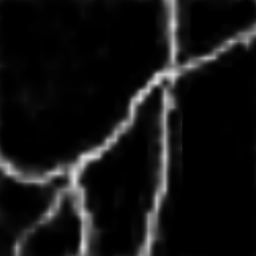}}
\quad
\subfloat[Scale-5,$P^{5}$]{\label{prob_5}\includegraphics[width=.18\textwidth]{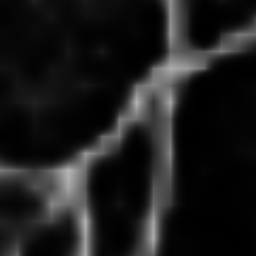}}
\caption{Visualization of Predicted Probability maps at each scale}
\label{prob_vis_fig}
\end{center}
\end{figure*}

\begin{figure*}[!h]
\begin{center}
\subfloat[Attn-1]{\label{att_1}\includegraphics[width=.18\textwidth]{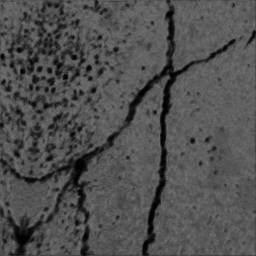}}
\qquad
\subfloat[Attn-2]{\label{att_2}\includegraphics[width=.18\textwidth]{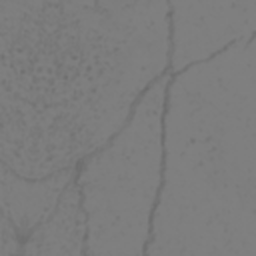}}
\qquad
\subfloat[Attn-3]{\label{att_3}\includegraphics[width=.18\textwidth]{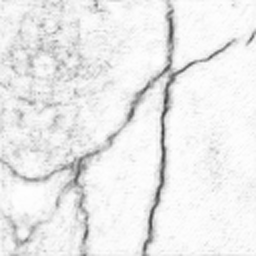}}
\qquad
\subfloat[Attn-4]{\label{att_4}\includegraphics[width=.18\textwidth]{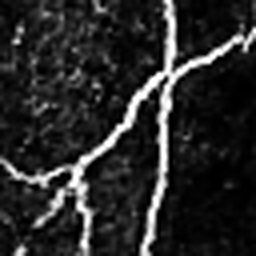}}
\qquad
\subfloat[Attn-5]{\label{att_5}\includegraphics[width=.18\textwidth]{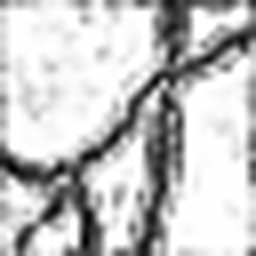}}
\qquad
\subfloat[Attn-6]{\label{att_6}\includegraphics[width=.18\textwidth]{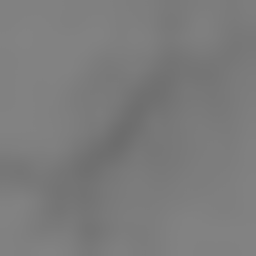}}
\caption{Visualization of Attention Maps in Encoder Scale-space. See SCNet
Architecture Diagram for Unit Names and Placements}
\label{attn_vis_fig}
\end{center}
\end{figure*}


Figures~\ref{pr_code}, \ref{pr_cfd}, \ref{pr_deep} and~\ref{pr_metu} show the \textit{Precison-Recall} curves of each model, on each dataset.
The figures also provide the AUPRC value for each model, per dataset. Given
the very small \% of foreground class pixels as baseline AUPRC per
dataset(refer Table \ref{data_imb}), it is obvious that though all models
have robust AUPRC for the foreground when compared to the baseline, our
model has the best AUPRC among all models, on \textit{each} dataset. Notably, for
same precision, we have better recall value among models. This in turn
can be attributed to reduction in missed detections in general, due to
usage of attention mechanism. Lesser missed detection is something that is
most sought after, in anomaly detection problems, where anomalies are rare.

\begin{table}[!h]
\begin{center}
\caption{Cross-datasets Performance of SCNet}
\label{cross_dataset_tab}
\begin{tabular}{|c|c|c|c|c|}
\hline
\multirow{2}{*}{\textbf{Trained On}} & \multicolumn{4}{c|}{\textbf{Tested On}}                                        \\ \cline{2-5} 
                                     & \textbf{Deepcrack} & \textbf{CrackForest} & \textbf{Codebrim} & \textbf{METU}  \\ \hline
\textbf{Deepcrack}                   & \textbf{91.23}     & 63.59                & 42.14             & 58.45          \\ \hline
\textbf{CrackForest}                 & 64.89              & \textbf{90.4}        & 42.38             & 57.2           \\ \hline
\textbf{Codebrim}                    & 63.08              & 63.64                & \textbf{64.22}    & 56.79          \\ \hline
\textbf{METU}                        & 65.78              & 64.48                & 51.32             & \textbf{69.13} \\ \hline
\end{tabular}
\end{center}
\end{table}

\subsection{Qualitative Performance}
The qualitative performance of each model, against a representative input belonging to each dataset, can be seen in Fig.~\ref{seg_res_fig}. Each column depicts the prediction output (including one column for ground truth), while each row names the dataset to which the each different input image belongs to. It can be seen that predictions of our model are more close to the GT, than other models.
\begin{figure*}[!t]
\begin{center}
\subfloat[Image]{\label{inp_img}\includegraphics[scale=.4]{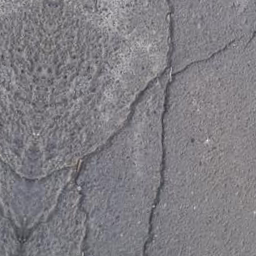}}
\qquad
\subfloat[Ground-Truth]{\label{inp_gt}\includegraphics[scale=.4]{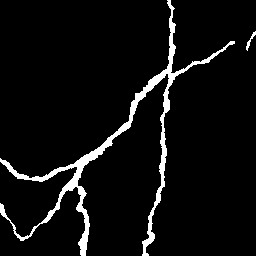}}
\qquad
\subfloat[Segmented Map]{\label{scnet_pred}\includegraphics[scale=.4]{img/qual/scnet_cfd_new.png}}
\qquad
\subfloat[Heatmap]{\label{pred_heatmap}\includegraphics[scale=.4]{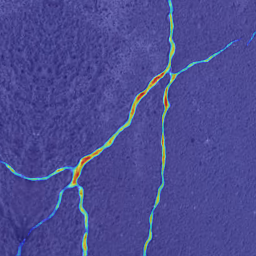}}
\caption{Grad-CAM Heatmap Visualization of Final Prediction}
\label{heatmap_fig}
\end{center}
\end{figure*}

\subsection{Discussion}

To understand how our model provides the required performance, we hereby
provide and explain a series of visualizations. The contribution of
individual scales in the final prediction has been shown in
Fig.~\ref{prob_vis_fig}, using probability maps $P^i$. It is clear from
this figure that all scales in the network try to capture crack-like
structures in the image. Further, the probability maps from the shallower
stages contribute more in the final prediction, as is expected and
explained in section \ref{attn_sec}. To understand deeper about performance at each scale, Fig.~\ref{attn_vis_fig} shows the
visualization of the attention maps learnt at each scale of the encoder, by
the corresponding attention module. Two important aspects become clear from
these attention maps. One, in lack of any obvious texture in background,
the attention module makes crack pixels provide contextual support among
themselves, to boost their classification probability. Two, the attention
maps generated by deeper encoder stages also produce sharp and significant
contextual attention, thereby being relevant to the model design even when
the corresponding probability maps were found by us to be diffused a bit,
as in Fig.~\ref{prob_vis_fig}. The variation in these attention maps
justifies the use of multiple attention modules, one at each scale, rather
than a single attention module at the end of the encoder, as is the case
with with architectures of many a popular attention-based segmentation
architectures, e.g. \cite{danet_pap, ccnet_pap}.

To test the low-data adaptability property of our model, we tested SCNet in a \textit{transfer learning} setup. Briefly, we trained the model on one dataset, and tested it on another. The results are tabulated in Table~\ref{cross_dataset_tab}. It can be seen that our model is able to learn generic features from various datasets, and transfer quite well on any other infrastructure dataset.

We also generate the segmentation heatmap using popular Grad-CAM visualization technique \cite{grad_cam_pap}, shown in Fig.~\ref{heatmap_fig}. It can be seen from this figure that as a \textit{limitation}, SCNet mostly fails to detect lighter cracks with strong confidence. This in turn leads to false negatives. Few times when thin cracks do get detected, there is additional detection of a few pixels beyond the actual width of the crack, resulting in the false positives as well.

\begin{figure*}[!h]
\begin{center}
\subfloat[Deepcrack]{\label{tp_dp}\includegraphics[scale=.39, trim=103 43 90 50, clip]{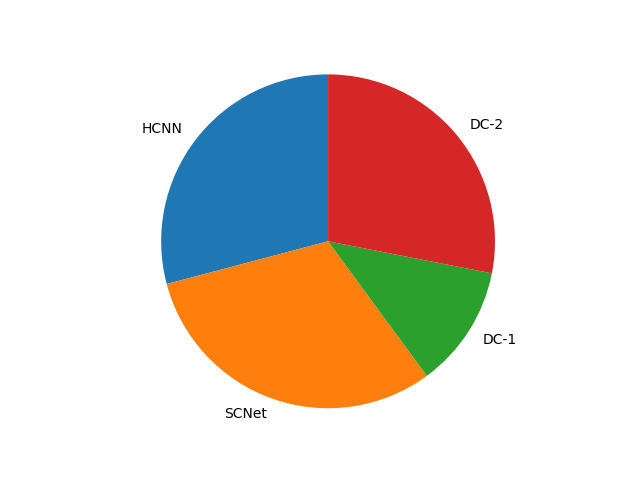}}
\qquad
\subfloat[CrackForest]{\label{tp_cfd}\includegraphics[scale=.39, trim=103 43 100 50, clip]{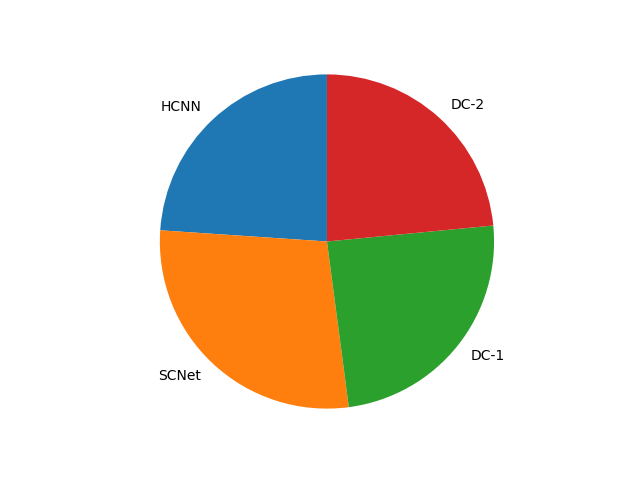}}
\qquad
\subfloat[CODEBRIM]{\label{to_code}\includegraphics[scale=.39, trim=103 43 80 50, clip]{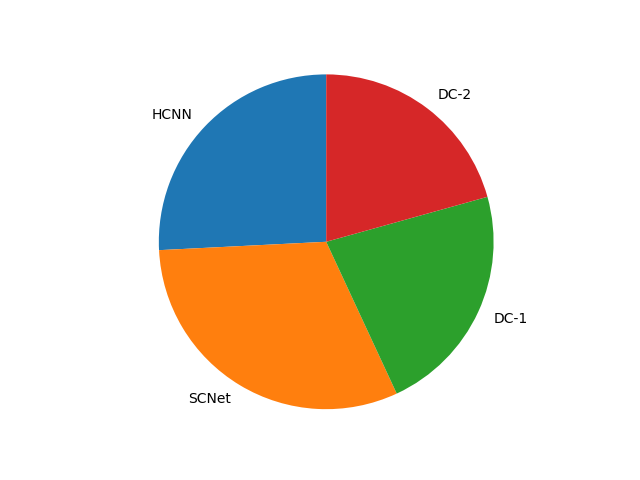}}
\qquad
\subfloat[METU]{\label{tp_metu}\includegraphics[scale=.39, trim=103 43 100 50, clip]{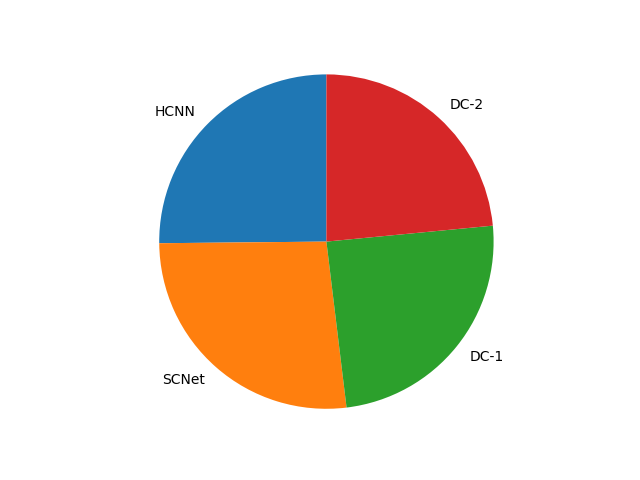}}
\caption{Visualization of True Positives for each model, on each dataset}
\label{tp_fig}
\end{center}
\end{figure*}

\begin{figure*}[!h]
\begin{center}
\subfloat[Deepcrack]{\label{fp_dp}\includegraphics[scale=.39, trim=81 40 103 50, clip]{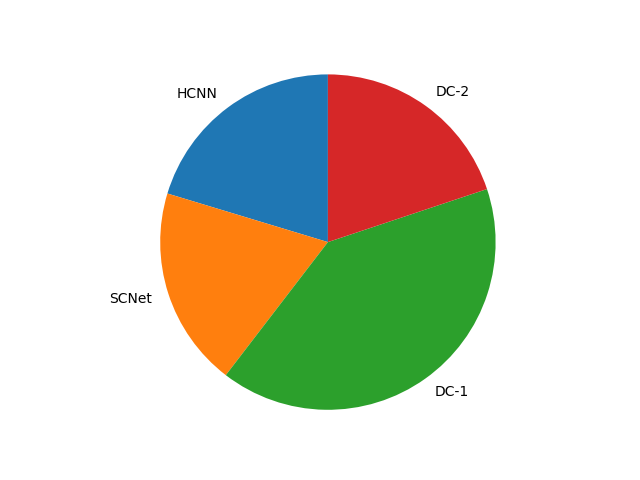}}
\qquad
\subfloat[CrackForest]{\label{fp_cfd}\includegraphics[scale=.39, trim=97 40 95 50, clip]{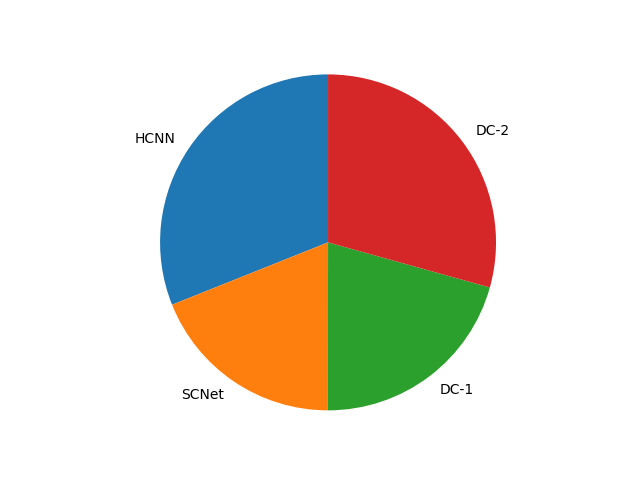}}
\qquad
\subfloat[CODEBRIM]{\label{fp_code}\includegraphics[scale=.39, trim=102 40 77 50, clip]{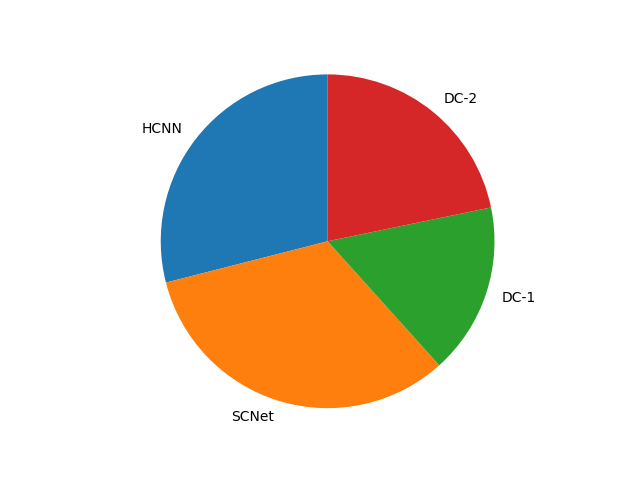}}
\qquad
\subfloat[METU]{\label{fp_metu}\includegraphics[scale=.39, trim=78 40 71 50, clip]{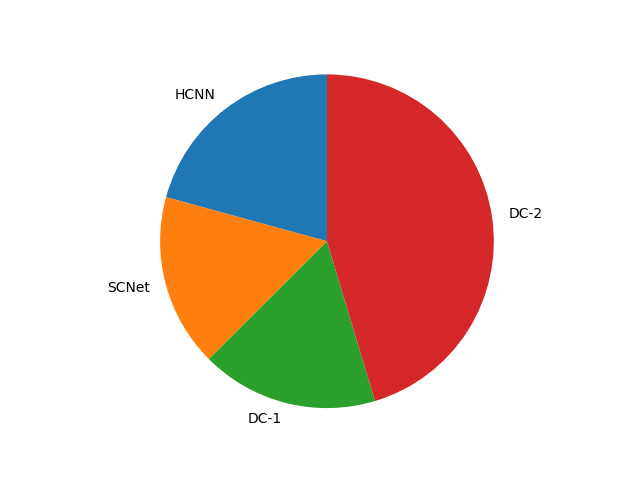}}
\caption{Visualization of False Positives for each model, on each dataset}
\label{fp_fig}
\end{center}
\end{figure*}

\begin{figure*}[!h]
\begin{center}
\subfloat[Deepcrack]{\label{fn_dp}\includegraphics[scale=.39, trim=90 35 103 45, clip]{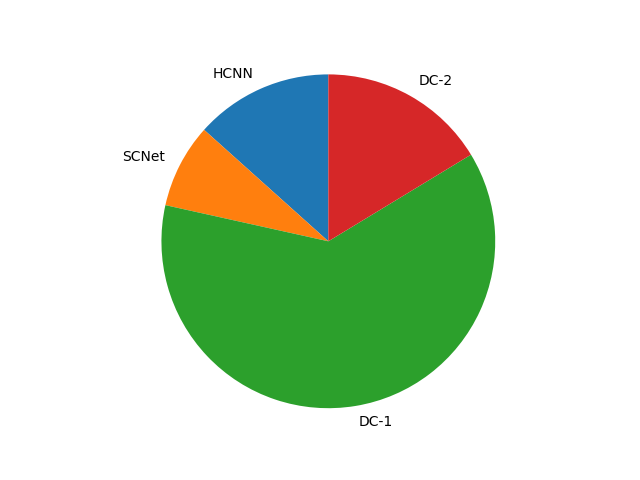}}
\qquad
\subfloat[CrackForest]{\label{fn_cfd}\includegraphics[scale=.39, trim=98 35 90 50, clip]{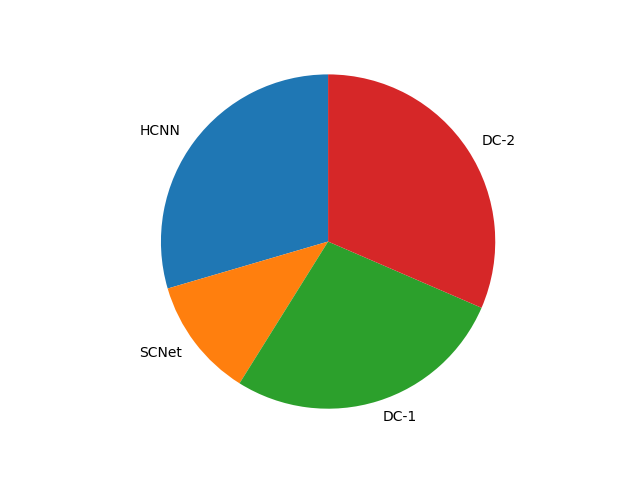}}
\qquad
\subfloat[CODEBRIM]{\label{fn_code}\includegraphics[scale=.39, trim=100 35 90 50, clip]{img/pie_charts/fn_cfd.png}}
\qquad
\subfloat[METU]{\label{fn_metu}\includegraphics[scale=.39, trim=105 35 100 50, clip]{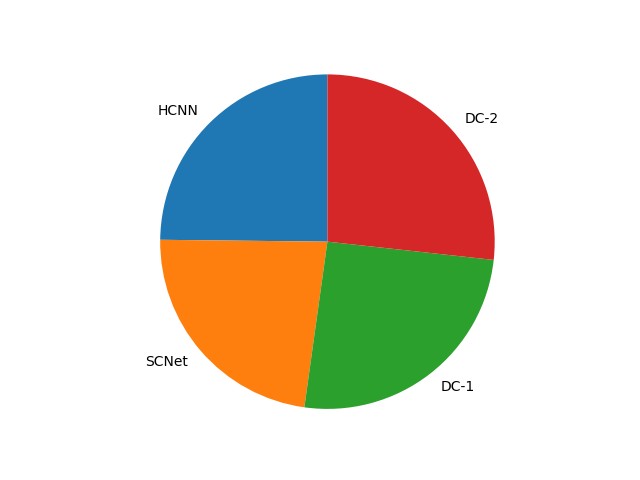}}
\caption{Visualization of False Negatives for each model, on each dataset}
\label{fn_fig}
\end{center}
\end{figure*}

As another visualization, we provide statistical trends on how our model
performs relatively better. Fig.~\ref{tp_fig} shows the relative \% of true
positives classified by the 4 models being compared, on a per-dataset
basis. It can be seen that our SCNet model classifies, for each of the
datasets, non-trivial amount of more true positive pixels, compared to
other 3 models. Similarly, Fig.~\ref{fp_fig} shows the relative \% of false positives detected by the same 4 models, on a per-dataset basis. Other than the exception of CODEBRIM dataset, for each of the remaining datasets, SCNet model classifies lesser number of false positive pixels, by a non-trivial amount. Finally, Fig.~\ref{fn_fig} shows the relative \% of false negatives/missed detections by the 4 models being compared, on a per-dataset basis. Yet again, it can be seen that on each of the datasets, SCNet model misses out on lesser number of true classifications, by a non-trivial amount. All these trends point to the importance of each of the various novel elements that we have brought out in our model design, and are intuitively explained in section \ref{model_sec}. The relative importance of these elements is established in the next section, via ablation studies.

\begin{figure*}
\begin{center}
\subfloat[]{\label{gap1}\includegraphics[scale=0.3]{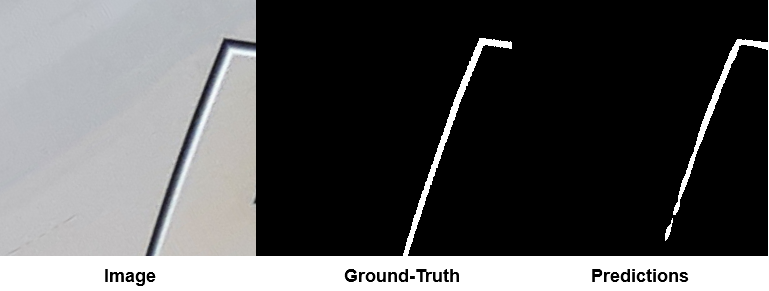}}
\qquad
\subfloat[]{\label{gap2}\includegraphics[scale=0.3]{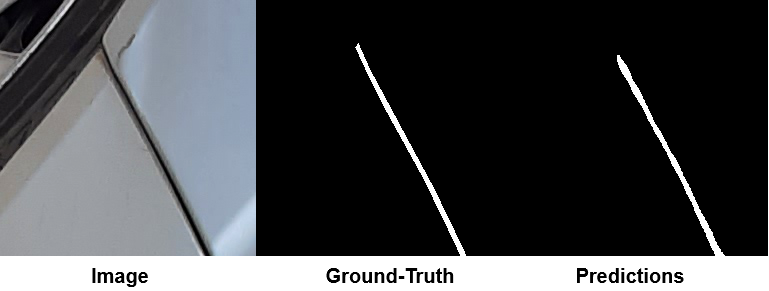}}
\caption{Illustrative Predictions of SCNet on Car Door Gaps}
\label{cargap_fig}
\end{center}
\end{figure*}

\section{Experiments and Results on Industrial Datasets}
\begin{figure}[!h]
\begin{center}
\includegraphics[width=\linewidth]{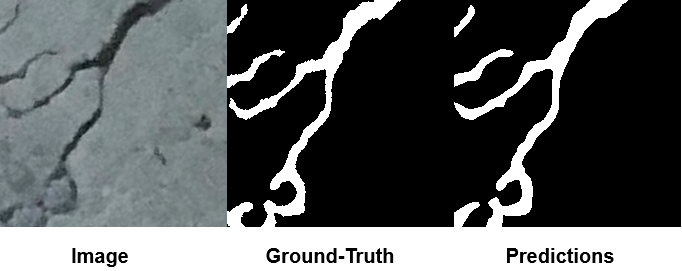}
\caption{Illustrative Prediction of SCNet on BMW Dataset}
\label{bmw_fig}
\end{center}
\end{figure}

\begin{figure}[!h]
\begin{center}
\includegraphics[width=\linewidth]{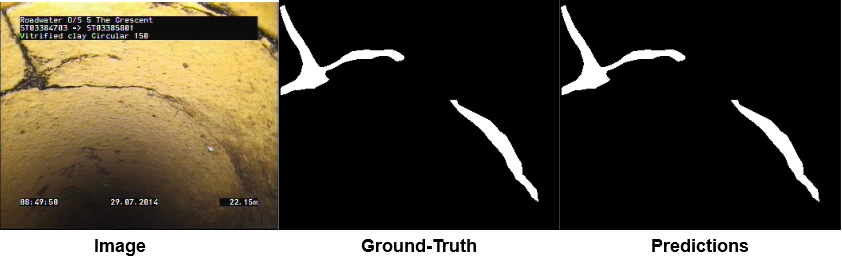}
\caption{Illustrative Prediction of SCNet on Wessex Water Dataset}
\label{ww_fig}
\end{center}
\end{figure}

We were able to further test our model on two challenge datasets provided by BMW and Wessex Water respectively. The former dataset was based on drone-based inspection of road networks, while the latter dataset was based on robotic inspection of interiors of a water supply pipeline/network. We did not benchmark other models (DC-1, HCNN and DC-2) on these datasets, but instead computed pixelwise F1 score and IoU score of SCNet. The results are tabulated in Tab.~\ref{ext_res_tab} and illustrated in Figures.~\ref{bmw_fig} and ~\ref{ww_fig}.

\begin{table}[!h]
\begin{center}
\caption{Performance on Industrial Datasets}
\label{ext_res_tab}
\begin{tabular}{|c|c|c|}
\hline
\textbf{Dataset} & \textbf{Pixelwise F1 score} & \textbf{IoU Score}  \\ \hline \hline
Wessex Water & 98.63 & 93.17 \\ \hline
BMW & 94.87 & 90.91 \\ \hline
\end{tabular}
\end{center}
\end{table}

\section{Experiments and Results on Non-Infrastructural Datasets}
Other than cracks in infrastructures, we were able to significantly generalize and prove the effectiveness of SCNet in two manufacturing scenarios.

\subsection{Car Door Gap Analysis}
The objective of this experiment was to check effectiveness of SCNet on predicting gaps between car doors, for quality inspections. The problem required the metric accuracy of the prediction in millimeters. The pixel-wise F1-score of SCNet for prediction came to be 95.77\%, and IoU score turned out to be 91.88\%, both of which are robust. Few illustrative predictions are shown in Fig.~\ref{cargap_fig}. 

\subsection{Steel Surfaces}
\begin{figure}[!h]
\begin{center}
\includegraphics[width=\linewidth]{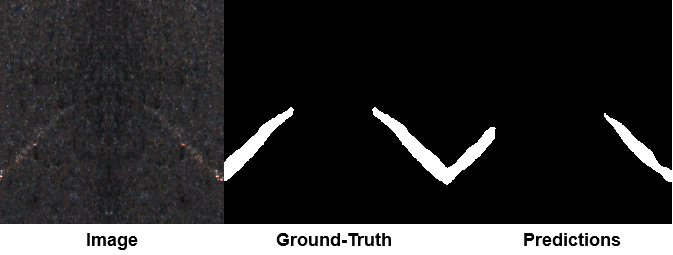}
\caption{Illustrative Prediction of SCNet on Multiple Steel Surface Defects}
\label{steel_fig}
\end{center}
\end{figure}

We also tested our model on steel surface defects as provided in KolektorSDD2 dataset \cite{kolektor_dset_pap}. Here, only scratches resemble cracks, while SCNet is employed to predict \textbf{more than 5 other defects} as well, present in the dataset. Cumulatively, the results are encouraging: we got a pixelwise F1 score of 77.29\% across all types of defects. Since defect classification is not available, we could not test for performance on segmentation of each defect in isolation. An illustrative prediction of steel surface defect is shown in Fig.~\ref{steel_fig}.

\begin{table*}[!t]
\caption[]{Ablation Study on Network Architecture\footnotemark}
\label{imprv_table}
\begin{tabular}{|c|c|c|c|c|c|c|c|c|c|c|c|c|c|c|}
\hline
\multirow{2}{*}{\textbf{Dataset}} & \multicolumn{2}{c|}{\textbf{Baseline}} & \multicolumn{2}{c|}{\textbf{Baseline+Attn}} & \multicolumn{2}{c|}{\textbf{\begin{tabular}[c]{@{}c@{}}Baseline+Attn\\ +Focal Loss\end{tabular}}} & \multicolumn{2}{c|}{\textbf{\begin{tabular}[c]{@{}c@{}}Baseline+Attn\\ +Focal Loss\\+Soft\_IoU\end{tabular}}} & \multicolumn{2}{c|}{\textbf{\begin{tabular}[c]{@{}c@{}}Baseline\\ +ScalarWeights\end{tabular}}} & \multicolumn{2}{c|}{\textbf{\begin{tabular}[c]{@{}c@{}}Baseline+Attn\\ +Focal+Soft-Iou\\ +Edges-1 level\end{tabular}}} & \multicolumn{2}{c|}{\textbf{\begin{tabular}[c]{@{}c@{}}Peak\\ Performance\end{tabular}}} \\ \cline{2-15} 
                                  & \textbf{F1}     & \textbf{Size}    & \textbf{F1}       & \textbf{Size}       & \textbf{F1}                                  & \textbf{Size}                                 & \textbf{F1}                                       & \textbf{Size}                                      & \textbf{F1}                                 & \textbf{Size}                                 & \textbf{F1}                                             & \textbf{Size}                                            & \textbf{F1}                                    & \textbf{Size}                                    \\ \hline
\textbf{Deepcrack}                & 87.56           & 29.46           & 89.27             & 29.46              & 89.98                                        & 29.46                                        & 90.57                                             & 29.46                                             & 86.91                                       & 29.46                                        & 91.91                                                   & 15.30                                                   & 91.23                                          & 29.46                                           \\ \hline
\textbf{CODEBRIM}                 & 57.92           & 29.46           & 62.65             & 29.46              & 63.28                                        & 29.46                                        & 63.60                                             & 29.46                                             & 57.79                                       & 29.46                                        & 62.69                                                   & 15.30                                                   & 64.22                                          & 29.46                                           \\ \hline
\textbf{CrackForest}              & 79.92           & 29.46           & 86.29             & 29.46              & 87.59                                        & 29.46                                        & 88.80                                             & 29.46                                             & 79.67                                       & 29.46                                        & 90.68                                                   & 15.30                                                   & 90.40                                          & 29.46                                           \\ \hline
\textbf{METU}                     & 65.40           & 29.46           & 67.54             & 29.46              & 68.26                                        & 29.46                                        & 68.57                                             & 29.46                                             & 65.62                                       & 29.46                                        & 66.95                                                   & 15.30                                                   & 69.13                                          & 29.46                                           \\ \hline
\end{tabular}
\end{table*}

\section{Ablation Studies and Design Options}
\label{abl_sec}
We conducted extensive ablation experiments with different focuses, to understand the relative importance of model elements. These experiments cover both the aspects of any machine learning model, namely structural variations and the variations over runtime dynamics. They cover all the improvements described in section \ref{model_sec}, to establish rationality of their usage and to understand the relative advantage that they provide. Few experiments have also been conducted, to evaluate various design options. The experiments have been performed on all datasets, instead of one, to make more robust inferences.

\subsection{Architecture-related Ablations}
As primary set of ablations, we studied the effect of adding/removing
various units in the SCNet architecture, on its performance. This set of variations introduces a tradeoff, between the performance and the number of learnable parameters (in millions). The latter impacts the model size, which is reported as well. The effects are summarized in Table \ref{imprv_table}. In the table, the baseline corresponds to HCNN implementation, while peak performance is based on our full model. Also, `ScalarWeights' corresponds to usage of single trainable scalar instead of a filter in the attention operator, `Edges' correspond to usage of 4$^{th}$ channel at input, while `-1 level' corresponds to 4-level scale space instead of 5-level. One can see from the table that
\begin{itemize}
    \item Among all novelties, introducing the spatial attention mechanism leads to most significant improvement in the segmentation performance, over the baseline \cite{hcnn_pap}, without much increase in the trainable parameters.
    \item Instead of using the attention module at various places, we also tried to use the trainable scalar weights for feature maps, at both the encoder and decoder side, but it did not improve the performance. 
    \item We also tried to reduce the network complexity, using $4$ scales instead of $5$. As is visible in 2$^{nd}$ last column, the performance decreased for the more challenging datasets.
    \item The use of a combination of focal loss and Soft-IoU loss gives modest improvements in the segmentation performance.
\end{itemize}

\begin{table}[!h]
\begin{center}
\caption{Effect of Different Normalization and Pooling}
\label{norm_pool}
\begin{tabular}{|c|c|c|}
\hline
\multirow{2}{*}{\textbf{Dataset}} & \textbf{Inst. Norm. +}     & \textbf{Inst. Norm. +}         \\
                         & \textbf{Max Pool} & \textbf{Dilated Conv} \\ \hline
Deepcrack                & 91.23             & 90.82                 \\ \hline
CODEBRIM                 & 64.22             & 62.01                 \\ \hline
METU                     & 69.13             & 66.65                 \\ \hline
CrackForest              & 90.4              & 88.64                 \\ \hline
\end{tabular}
\end{center}
\end{table}

\subsection{Loss/Optimization-related Ablations}
In a complementary set of ablation experiments, we kept the architecture intact, but varied and benchmarked the dynamic behavior of the model.
We first tried to initialize the model weights in different ways, to study transfer learning. We tried with standard Xavier initialization, initialization via training on a different crack dataset, and a curious (industrial) dataset having samples of surfaces and textures of various materials in everyday usage \cite{minc_pap}. It can be seen from Table~\ref{weight_tab} Xavier initialization performed the best. This implies that as another limitation, our model is currently not able to handle domain shift.

\begin{table}[!h]
\begin{center}
\caption{Effect of Different Weight Initializations}
\label{weight_tab}
\setlength{\tabcolsep}{0.5em} 
{\renewcommand{\arraystretch}{1.2}
\begin{tabular}{|c|c|c|c|}
\hline
\multirow{2}{*}{\textbf{Dataset}} & \multirow{2}{*}{\textbf{Xavier}} & \multirow{2}{*}{\textbf{Crack260}} & \textbf{Encoder with}        \\ 
                                  &                                  &                                    & \textbf{MINC\cite{minc_pap}} \\ \hline
Deepcrack                         & \textbf{91.23}                   & 90.45                              & 86.32                        \\ \hline
CODEBRIM                          & \textbf{64.22}                   & 62.54                              & 49.35                        \\ \hline
METU                              & \textbf{69.13}                   & 66.32                              & 64.21                        \\ \hline
CrackForest                       & \textbf{90.4}                    & 88.89                              & 76.52                        \\ \hline
\end{tabular}
}
\end{center}
\end{table}

We next tried to vary the loss combinations under multi-task learning
setup. We additionally tried training the model with binary cross-entropy loss as well as Lov\'asz loss, designed for pixel-wise regression \cite{lovasz_loss_pap}. While it becomes obvious from Table~\ref{loss_tab} that focal loss along with Soft-IoU loss gave best performance, the impact of using soft-IoU loss additionally is modest. This can can perhaps be attributed to coarser, region-level supervision provided by soft-IoU loss, as against fine-grained, pixel-level supervision provided by focal loss. As an aside, we also tried replacing max-pool with dilated convolutions, as is a popular practice today. But as Table~\ref{norm_pool} suggests, it did not yield any improvement.

\begin{table}[!h]
\begin{center}
\caption{Effect of Different Loss Formulations}
\label{loss_tab}
\setlength{\tabcolsep}{0.5em} 
{\renewcommand{\arraystretch}{1.2}
\begin{tabular}{|c|c|c|c|c|}
\hline
\multirow{2}{*}{\textbf{Dataset}} & \textbf{\textbf{Peak}} & \textbf{\textbf{CE Loss}} & \textbf{CE Loss +}     & \textbf{Focal Loss +}  \\
                                  & \textbf{Performance}   & \textbf{Only}             & \textbf{Soft-IoU loss} & \textbf{Lov\'asz Loss} \\ \hline
\textbf{Deepcrack}                & \textbf{91.23}         & 90.53                     & 90.73                  & 90.15                  \\ \hline
\textbf{CODEBRIM}                 & \textbf{64.22}         & 62.83                     & 62.87                  & 63.28                  \\ \hline
\textbf{METU}                     & \textbf{69.13}         & 68.61                     & 67.84                  & 67.13                  \\ \hline
\textbf{CrackForest}              & \textbf{90.40}         & 88.44                     & 88.57                  & 86.99                  \\ \hline
\end{tabular}
}
\end{center}
\end{table}

\footnotetext{Model size is in millions of parameters}

\section{Conclusion}
In this paper, we have introduced a model for pixel-wise segmentation of
crack faults in infrastructures. Crack faults are one of the most critical
and common faults, that can lead to failure of infrastructural systems and
catastrophe. We have attempted to tackle the dual challenge of no obvious
pattern across crack foreground regions, against variable and poor-textured
backgrounds. We use multiple design novelties, with usage of scale-space
attention in both encoder and decoder giving us the best improvement in
performance. We have shown the consistency of improvement, and widespread
applicability of our model, by rigorously evaluating our model against 4
infrastructure datasets of different natures and sizes. In all cases, our model
outperforms the state-of-the-art by a significant margin, in terms of both
F1 score and AUPRC. It was even successfully deployed and tested over two industry-provided datasets, collected in challenging conditions. Our performance improvement is without any careful tuning of hyperparameters, or post-processing. Post such enhancements, the
performance of our algorithm is only bound to increase. Our work thus
establishes a new baseline for the crack segmentation task. As a curious benchmarking, we could also successfully demonstrate our model's performance in two manufacturing inspection scenarios, on vastly different anomalies. In future, to further expand its scope in manufacturing applications, we
intend to more rigorously benchmark it on wide variety of \textit{material quality inspections}, such as cracks, wrinkles, folds and scratches on leather hide, rexine cover, metal sheet etc. 
We are hopeful that in SCNet, we will thus be able to establish a truly generic defect segmentation model.

\section*{Acknowledgment}
The authors gratefully acknowledge the guidance of Dr. Hiranmay Ghosh at the starting of this research work. 

\ifCLASSOPTIONcaptionsoff
  \newpage
\fi

\bibliography{ms}

\end{document}